\documentclass[10pt,twocolumn,letterpaper]{article}

\usepackage{preprint}
\usepackage{placeins}

\usepackage{multirow}
\usepackage{makecell}
\usepackage{colortbl}
\usepackage{tcolorbox}
\usepackage{float}
\usepackage{cuted}

\definecolor{blkone}{HTML}{EAF1FA}
\definecolor{blktwo}{HTML}{D3E3F3}
\newcommand{\cmark}{\ensuremath{\checkmark}}
\newcommand{\xmark}{\ensuremath{\times}}
\newcommand{\rgt}{r^{\mathrm{gt}}}

\renewcommand{\paragraph}[1]{\vspace{.35em}\noindent\textbf{#1}\hspace{.4em}}

\definecolor{linkblue}{rgb}{0.21,0.49,0.74}
\usepackage[breaklinks,colorlinks,allcolors=linkblue]{hyperref}

\newcommand{\supp}[1]{App.~\ref{#1}}

\title{What Transfers from Judging Grasps? Graded Action Assessment\\as Pre-Training for Vision-Language-Action Models}

\author{Jayaraman J. Thiagarajan \qquad Rushil Anirudh \qquad Bharathan Balaji\\[0.3em]
{\normalsize Amazon AGI}}

\begin{document}
\maketitle

\begin{abstract}
Vision-language-action (VLA) policies inherit their visual and spatial understanding from a pre-trained
vision-language backbone, yet further pre-training with descriptive robot objectives such as embodied visual question
answering yields little downstream benefit. We ask whether an \emph{evaluative} objective, judging which of several
nearby candidate actions is physically viable, transfers better. We introduce Action Assessment Pre-Training (AAPT),
which trains a VLM to assess a scene and a candidate grasp, given as a 6-DoF pose and a gripper width, generate a physical reasoning trace about the
contact geometry, and output a graded viability score. Candidates are synthesized around annotated successes, so many
comparisons involve near-misses that differ by millimeters or degrees. After pre-training, the assessment interface
is discarded and only the backbone initializes the policy. Controls on identical candidates and matched seeds probe
what transfers. Graded targets beat a predictive control supervised by simulated collision and stability outcomes on
every seed, a ranking reward adds a gain that compute-matched continued training does not reproduce, and breaking the
correspondence between a candidate and either its trace or its score removes the gain in seed-1 alignment diagnostics.
Across two VLA pipelines, AAPT
reaches $88.8\%$ on zero-shot LIBERO-Plus and $4.54$ on CALVIN ABC$\rightarrow$D, gains most under camera and sensor
perturbations, and with half of the demonstrations already exceeds a base policy trained on all of them.
\end{abstract}

\section{Introduction}
\label{sec:intro}

Vision-language-action models attach an action decoder to a pre-trained vision-language
backbone~\cite{rt2, openvla, pi0, pi05}. The backbone supplies the policy's visual and spatial understanding of the
scene, so a natural way to improve policies is to improve the backbone before any action training. The evidence for
this is mixed. Scaling the backbone does not monotonically improve manipulation (\supp{app:scale}),
\citet{turbovla} obtain competitive results with no language backbone at all, and further pre-training on embodied
visual question answering gives little downstream benefit~\cite{vlm4vla}.

\begin{figure}[t]
\centering
\includegraphics[width=\linewidth]{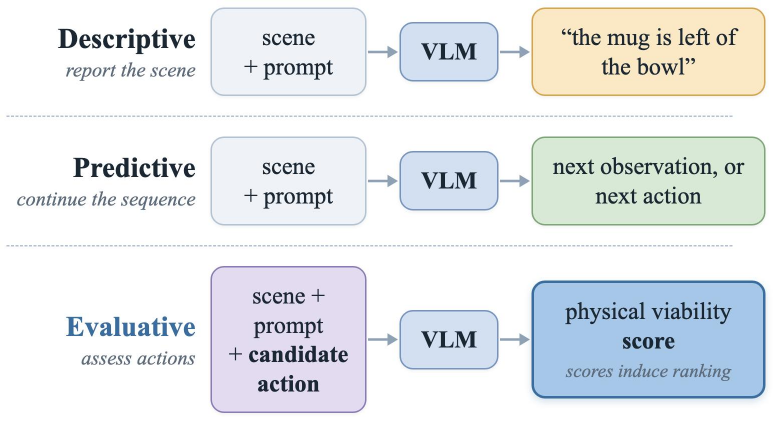}
\caption{Three families of pre-training objectives. Descriptive and predictive objectives never ask the backbone to
compare nearby actions. We study whether an evaluative objective, which ranks candidate actions by physical
viability, transfers better to control.}
\label{fig:taxonomy}
\end{figure}

We argue that the type of objective matters (\cref{fig:taxonomy}). \emph{Descriptive} objectives report what is
present, as in captioning and visual QA~\cite{llava15, spatialvla, robo2vlm, vqasynth}. \emph{Predictive} objectives
produce a continuation, such as a future frame, a world-model rollout or the next action~\cite{lapa, dreamvla,
mmada-vla}. Neither asks the model to compare nearby actions and decide which one will work. An \emph{evaluative}
objective does exactly that.

Judging a candidate grasp is a demanding visual task. The model has to relate a 6-DoF end-effector pose to the
object's geometry as it appears in the image, including the approach direction, where the fingers make contact and
whether the gripper opening fits the object. Near a successful grasp these quantities change by a few millimeters or
degrees, so object identity and appearance alone cannot decide the answer. A manipulation policy needs the same skill
whenever it brings the end effector into contact with an object, whether it grasps, pushes, slides or opens.

\begin{figure*}[t]
\centering
\includegraphics[width=0.86\textwidth]{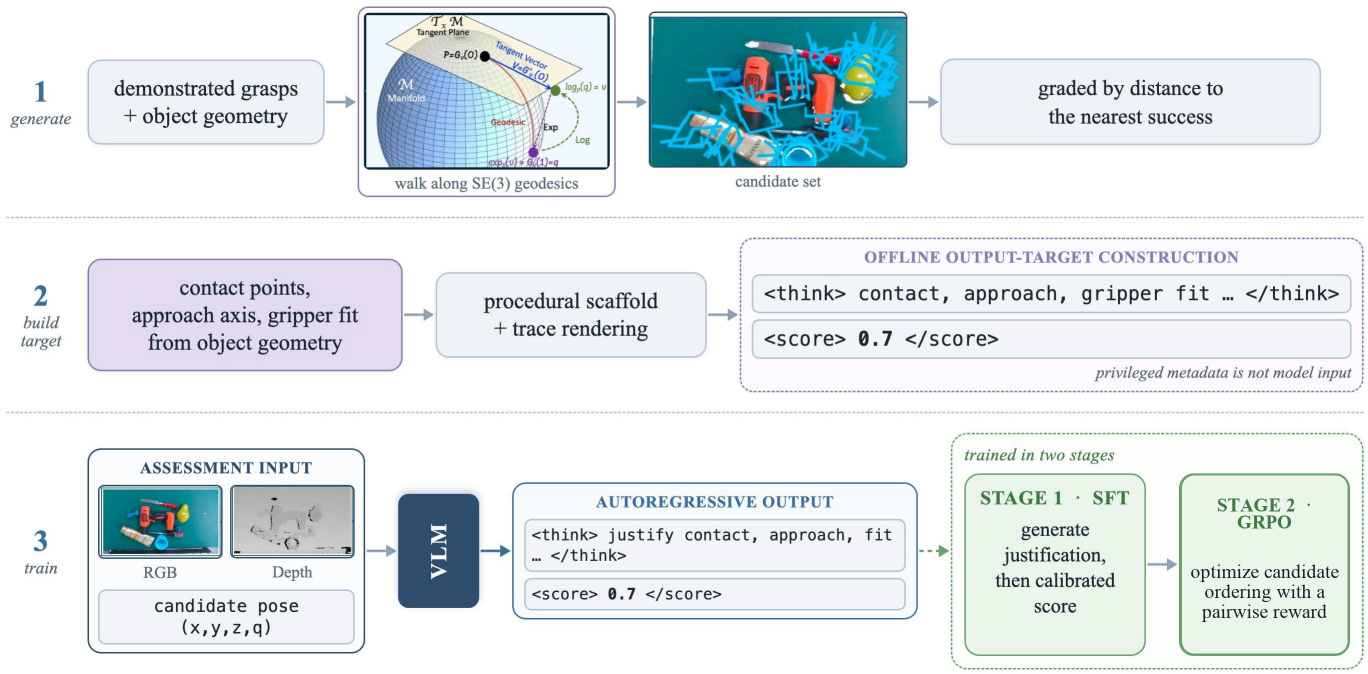}
\caption{Action Assessment Pre-Training. Perturbed grasps are graded by proximity to the nearest annotated success.
Given RGB, optional depth and one candidate grasp (pose and gripper width), the VLM generates a physical
reasoning trace followed by a score.
Privileged geometry constructs the targets but is never a model input. SFT calibrates the scores, and GRPO with a
ranking reward sharpens their ordering. Only the backbone weights transfer to the policy.}
\label{fig:aapt}
\end{figure*}

We introduce \textbf{Action Assessment Pre-Training (AAPT)} (\cref{fig:aapt}). Around each annotated successful
grasp, AAPT synthesizes a graded continuum of candidates, from near-misses at the success boundary to clear failures,
and grades each by its distance to the nearest known success. Given the scene and one candidate grasp (6-DoF pose and gripper width), the VLM first
generates a physical reasoning trace and then a viability score. Supervised fine-tuning calibrates the score, and GRPO
with a pairwise ranking reward sharpens the ordering of nearby candidates. After pre-training, the assessment
interface is discarded and only the backbone initializes the policy. We study grasping as a tractable first case,
because grasp viability can be judged from the scene and pose without credit assignment over a trajectory.

A gain from such pre-training has several possible explanations. The backbone might benefit simply from seeing
failures, from the extra optimization, or from learning to describe geometry, independently of graded evaluation. We
therefore build controls that match data, inputs and training seeds while varying the key forms of supervision.
A predictive control supervised by simulated collision and stability outcomes
on the same candidates helps, but graded targets help more on every seed. The ranking reward adds a gain that neither
compute-matched continued training nor a calibration-only reward reproduces. Breaking the correspondence between a
candidate and either its trace or its score removes the gain in seed-1 diagnostics, and breaking both makes the
policy worse than no pre-training. Together, the results point to \emph{candidate-aligned graded ordering} as the useful signal, beyond
failure exposure or additional optimization alone. Our contributions are as follows.
\begin{itemize}[leftmargin=1.2em, itemsep=1pt, topsep=2pt, parsep=0pt]
\item \textbf{An evaluative pre-training objective.} AAPT ranks near-miss grasp candidates through physically grounded
reasoning and transfers only the backbone to the policy.
\item \textbf{A controlled analysis of what transfers.} A same-data predictive control, paired-seed contrasts,
compute-matched controls and separate trace and score shuffles test the role of graded, candidate-aligned ordering.
\item \textbf{Transfer across pipelines and benchmarks.} AAPT improves a minimal~\cite{vlm4vla} and a
strong~\cite{acot-vla} VLA pipeline on LIBERO, LIBERO-Plus, CALVIN and SimplerEnv. Gains are largest under viewpoint
and sensor perturbations, on non-grasping and long-horizon tasks, and when demonstrations are scarce.
\item \textbf{Physically faithful assessment traces.} The pre-trained model's own traces match the ground truth on
$82\%$ of the fields that depend on the object's pose and geometry.
\end{itemize}

\section{Related work}
\label{sec:related}

\paragraph{Vision-language-action models.}
Most policies place a pre-trained VLM between observation and action. Autoregressive designs emit discretized action
tokens~\cite{rt2, openvla}, while continuous designs attach a regression head, flow-matching expert or diffusion
decoder~\cite{pi0, pi05, openvla-oft, vla-adapter}. Others drop the language model~\cite{turbovla} or the VLM
entirely~\cite{diffusion-policy, octo}. We study how the backbone is trained before an action head is attached.

\paragraph{Pre-training objectives for control.}
Spatial grounding adds metric and relational structure~\cite{spatialvla}, and embodied question answering turns
teleoperation data or synthetic scenes into visual QA pairs~\cite{robo2vlm, vqasynth}. Both are descriptive,
and~\citet{vlm4vla} report minimal downstream benefit. Predictive objectives generate a sub-goal
image~\cite{cot-vla}, an observation-action rollout~\cite{worldvla}, an enriched future
observation~\cite{dreamvla}, inverse dynamics~\cite{seer, gr1}, a latent action vocabulary~\cite{lapa, univla} or a
joint observation-action denoising~\cite{mmada-vla}. Their gains often come with a jointly trained action pathway
rather than the backbone alone. Prediction can also consume failed candidates, and our collision-and-stability
control tests this case on the same data as AAPT.

\paragraph{Evaluating actions.}
Robotics has long evaluated actions through grasp-quality networks such as Contact-GraspNet~\cite{contact-graspnet},
affordances~\cite{where2act}, action values~\cite{qtopt} and energies~\cite{ibc}. VIP~\cite{vip} and LIV~\cite{liv}
learn value-oriented representations from action-free video. AAPT instead conditions on an explicit candidate grasp
pose and learns the local ordering of successes, near-misses and failures. Existing evaluators
also stay active during
learning or deployment. Reward models supervise policy learning~\cite{interactive-reward}, and V-GPS~\cite{vgps} and
RoVer~\cite{rover} rerank actions at test time. AAPT discards the assessor and transfers only the backbone, so it is
orthogonal to inference-time reasoning in language~\cite{vlm2vla}, image space~\cite{cot-vla} or action
space~\cite{acot-vla}. Concurrently,~\citet{fact} train a world model on failure rollouts and mask the imitation
loss on failed actions. AAPT instead synthesizes graded alternatives around successes without collecting rollouts.

\section{Action Assessment Pre-Training}
\label{sec:method}

AAPT trains the backbone to distinguish viable grasps from near-misses and failures and to justify each judgment in
terms of contact geometry. \cref{tab:whosees} separates what each component sees. \supp{app:notation} lists all
symbols.

\begin{table}[t]
\centering
\footnotesize
\setlength{\tabcolsep}{3pt}
\begin{tabular}{@{}p{0.27\linewidth}p{0.69\linewidth}@{}}
\toprule
\textbf{Component} & \textbf{Content} \\
\midrule
Teacher only (offline) & Object mesh and pose, procedural scaffold and target reward
$\rgt$. They construct the output target and are never shown to the VLM. \\
VLM input $x$ & Scene RGB $I_{\mathrm{RGB}}$, optional depth $I_{\mathrm{depth}}$, candidate grasp $g$ as text
(position, quaternion and gripper width). \\
VLM output $y$ & Physical reasoning trace $z_{\mathrm{phys}}$, then \texttt{<score>} and a scalar $s\in[0,1]$. The
trace is an output, never an input. \\
Transferred & Backbone weights only. The prompt, trace and score interface are discarded. \\
\bottomrule
\end{tabular}
\caption{What each component sees during AAPT.}
\label{tab:whosees}
\end{table}

\subsection{Graded candidate sets}
Each object contributes annotated successful grasps, $\mathrm{SE}(3)$ perturbations near the success boundary and
clear failures, drawn from HouseCat6D~\cite{housecat6d} and GraspNet-1Billion~\cite{graspnet1b}. Each training set
holds $20$ candidates, one annotated success, one perturbation at each of four noise levels ($0.02$ to $0.12$ rad and
m) and $15$ dataset or uniformly sampled failures (\supp{app:candidates}). Binary labels would make a near-miss
indistinguishable from a gross failure, so we grade each candidate by its distance to the nearest known success. For
poses $\mathbf{T}=(\mathbf{R},\mathbf{t})$,
\begin{align}
d_{\tau}(\mathbf{T}_1,\mathbf{T}_2) &= \frac{\theta(\mathbf{R}_1,\mathbf{R}_2)}{\tau_R}
+ \frac{\lVert\mathbf{t}_1-\mathbf{t}_2\rVert_2}{\tau_t}, \nonumber\\
\rgt &= \frac{1}{1+d_{\min}/\tilde d},
\label{eq:reward}
\end{align}
where $\theta$ is the geodesic rotation angle, minimized over the parallel-jaw gripper's $180^\circ$ symmetry about
its approach axis so that physically identical grasps have zero distance, $\tau_R=0.02$\,rad, $\tau_t=0.02$\,m, $d_{\min}$ is the distance to
the nearest annotated success in the scene and $\tilde d$ is the median distance of that scene's failures. The
minimum accommodates multiple valid grasps, and the median normalizes scene scale. $\rgt$ is a relative geometric
target, not a calibrated success probability. Perturbations keep the seed's gripper width, so width never varies at
a fixed pose and $\rgt$ is a function of pose alone. The trace's gripper-fit statement describes the given opening
and does not enter the score. \cref{sec:target} validates $\rgt$ against simulated grasp execution.

\subsection{Physical reasoning traces}
To ground the score in contact geometry, the VLM generates a short physical reasoning trace before every score. The
target trace is built offline in two steps (\supp{app:traces}). A deterministic scaffold first computes, from the
object mesh and pose, the object's shape class and extents, its pose relative to the camera, the gripper's approach
direction in the camera and object frames, the grasp type, the contact height and lateral offset from the object's
axis, and the fit between gripper opening and the object's thinnest axis. Its qualitative descriptors, such as an
opening that is too narrow, well matched or too wide, are fixed geometric thresholds and do not depend on $\rgt$.
Qwen3-VL-8B then rewrites the scaffold as concise text $z_{\mathrm{phys}}$ that covers the object and its pose, the
approach and grasp type, and the contact and fit. The rewriting teacher also sees $\rgt$ and is asked to describe
evidence consistent with it, so its wording may carry target information. The trace never states the score, a
quality tier or any verdict. We call $z_{\mathrm{phys}}$ a physical reasoning trace for brevity. It is a distilled
description of privileged geometry, and we do not claim that it is a faithful reasoning process. \cref{fig:traceex} shows the scaffold content for
two candidates on the same object. Because the object description is shared, only the candidate-specific fields can
separate them. Because the trace precedes the score, the score is conditioned on the generated physical description
rather than followed by a post hoc explanation.

\begin{figure}[t]
\begin{tcolorbox}[colback=gray!5,colframe=gray!40,boxsep=1pt,left=3pt,right=3pt,top=2pt,bottom=2pt]
\scriptsize
\textbf{Shared.} Peach, roughly spherical, $6.2\times6.3\times5.9$\,cm, tilted, $0.48$\,m from the camera.\\[2pt]
\textbf{Highest-reward candidate.} Top-down approach along the longest axis, angled grasp near the top, $55$\,mm
off-center. Opening $6.3$\,cm, well matched to the thinnest axis ($5.9$\,cm), inserted $30$\,mm.\\[2pt]
\textbf{Lowest-reward candidate.} Approach from below or rear along the intermediate axis, side grasp near the top,
$28$\,mm off-center. Opening $9.4$\,cm, reasonable for the thinnest axis, inserted $10$\,mm.
\end{tcolorbox}
\caption{Procedural scaffold content for two candidates on one object (abridged). It is used offline to build the
target trace and never appears in the VLM prompt. The full scaffold and rendering prompts are in \supp{app:traces}.}
\label{fig:traceex}
\end{figure}

\subsection{Two-stage training}
\paragraph{Stage 1, calibration.} Supervised fine-tuning trains the language model (vision encoder frozen) to
generate $y=[z_{\mathrm{phys}},\texttt{<score>},\rgt]$ from $x$ by next-token prediction, for one epoch.

\paragraph{Stage 2, ordering.} Pointwise calibration does not directly optimize the ordering of nearby candidates,
so we refine it with GRPO~\cite{visualquality-r1, visual-rft}. For each candidate $g_i$, GRPO samples a group of
five completions from the same prompt. We write $s_i$ for the score parsed from one rollout and $\bar s_i$,
$\sigma_i$ for the mean and standard deviation of the group's scores. Rewards are normalized within each group. Each
rollout of $g_i$ is compared with the $K$ candidates nearest to $g_i$ in ground-truth rank under a
Thurstone-Gaussian model~\cite{thurstone1927, visualquality-r1},
\begin{align}
p_{\mathrm{gt}}(g_i\succ g_j) &= \sigma\!\big(\kappa(\rgt_i-\rgt_j)\big), \nonumber\\
p_{\mathrm{pred}}(g_i\succ g_j) &= \Phi\!\big((s_i-\bar s_j)/\sigma_{\mathrm{comb}}\big),
\end{align}
where $\Phi$ is the standard normal CDF and
$\sigma_{\mathrm{comb}}=\max(\sqrt{\sigma_i^2+\sigma_j^2+\gamma},\lambda)$. The pairwise reward
$r_{\mathrm{pair}}=\max\big(0,\,2[p_{\mathrm{gt}}p_{\mathrm{pred}}+(1-p_{\mathrm{gt}})(1-p_{\mathrm{pred}})]-1\big)$
measures agreement between the two preferences, and the rollout reward
\begin{equation}
R(g_i) = \alpha\,R_{\mathrm{pair}}(g_i) + (1-\alpha)\big(1-\lvert s_i-\rgt_i\rvert\big)
\label{eq:grpo}
\end{equation}
balances the mean pairwise reward $R_{\mathrm{pair}}$ over the $K$ comparisons with calibration. Restricting
comparisons to nearby ranks focuses updates on ambiguous candidates around the success boundary. We use $K=4$ and
$\alpha=0.6$ for 500 steps (\supp{app:optimization}).

\subsection{Transfer to a control policy}
The AAPT checkpoint replaces the base VLM in two pipelines (Supp.~Fig.~\ref{fig:pipelines}). VLM4VLA~\cite{vlm4vla} appends a
single action query token decoded by an MLP, so fewer than $1\%$ of its parameters are new and it is a direct test of
the backbone. ACoT-VLA~\cite{acot-vla} adds explicit and implicit action reasoners that read the backbone's key-value
cache, and tests whether the gain survives in a strong architecture. In both, \emph{all} policy parameters, including
the vision encoder, are then trained on the same demonstrations. During AAPT itself the vision encoder is frozen, so
the changed initialization lies in the multimodal language-model layers that process the scene. It is not a frozen
feature extractor, and the candidate-grasp interface it was trained with is discarded.

\begin{figure*}[t]
\centering
\includegraphics[width=0.92\textwidth]{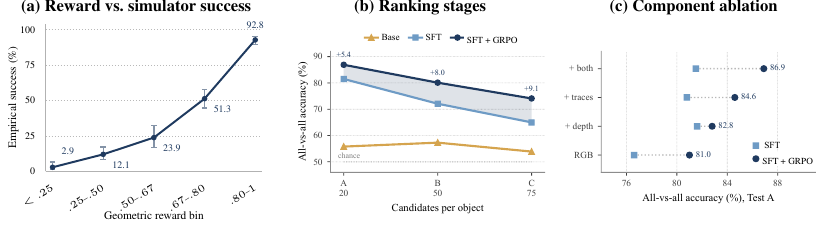}
\caption{Assessment diagnostics. \textbf{(a)} Fixed geometric target $\rgt$ versus empirical grasp success over
$1000$ stratified simulator executions (no model involved). \textbf{(b)} Pairwise ranking accuracy on $200$ held-out
scenes after SFT and GRPO as candidate sets grow from $20$ to $75$. \textbf{(c)} Component ablation on Test A. Full
metrics are in \supp{app:ranking}.}
\label{fig:ranking}
\end{figure*}

\section{Experimental setup}
\label{sec:exp}

\paragraph{Protocol.} Unless stated otherwise, AAPT uses Qwen3-VL-4B and a $500$K-example corpus. Base and AAPT
checkpoints initialize the same architecture and are trained on the same demonstrations with the same three training
seeds and data order, so the controlled difference is the initialization. Because matched runs share seed effects,
we report paired within-seed differences as well as means.

\paragraph{Assessment evaluation.} We hold out $200$ scenes and build three candidate sets per object instance. Test
A has $20$ candidates ($5$ annotated successes), Test B $50$ ($20$) and Test C $75$ ($25$), so larger sets contain
more plausible alternatives. We report all-versus-all pairwise ranking accuracy, the fraction of candidate pairs
whose predicted order agrees with $\rgt$ (chance $50\%$), and give Top-$M$ accuracy in \supp{app:ranking}.
The split is scene-disjoint but not object-disjoint, so it measures generalization to unseen scenes.

\paragraph{Implementation.} Stage 1 updates the language model at full rank for one epoch (learning rate
$5\times10^{-6}$, batch size $32$, cosine decay, $10\%$ warmup). Stage 2 runs $500$ GRPO steps at $3\times10^{-6}$
with five rollouts per candidate, $\kappa=8$, $\gamma=0.01$, $\lambda=0.25$, $K=4$, $\alpha=0.6$ and a KL penalty of
$5\times10^{-3}$. We transfer the final checkpoint without selecting on downstream performance. ACoT-VLA trains for $60$K
steps (learning rate $5\times10^{-5}$, batch size $128$). VLM4VLA trains for $50$K steps at batch size $512$ on LIBERO
and SimplerEnv and $60$K steps at batch size $128$ on CALVIN. Both pipelines see RGB and language only, so depth is
never available to the policy.

\paragraph{Benchmarks.} LIBERO~\cite{libero} measures multi-task success on four suites. LIBERO-Plus~\cite{libero-plus}
evaluates the same LIBERO checkpoints zero-shot under seven held-out perturbation factors. CALVIN
ABC$\rightarrow$D~\cite{calvin} measures the number of five chained instructions completed in a held-out environment.
SimplerEnv~\cite{simplerenv} provides a second-embodiment check. Protocols are in
\supp{app:downstream-protocols}. No object, scene or embodiment of these benchmarks appears in the AAPT corpus, so
downstream transfer is always to unseen objects and environments.

\paragraph{Controls.} All controls use the ACoT-VLA recipe. The matched progression fixes scenes, candidates and
inputs while changing the principal quality supervision; target-specific output formats are detailed below.
\begin{itemize}[leftmargin=1.1em, itemsep=1pt, topsep=1pt, parsep=0pt]
\item \emph{Collision and stability prediction} uses the same scenes, candidates, inputs and SFT schedule but predicts
two binary simulator outcomes, measured for every candidate. Its trace generator sees the candidate's geometry and
the two outcomes only, never $\rgt$ (Supp.~Fig.~\ref{fig:traceprompt-control}).
\item \emph{Graded SFT} is AAPT without Stage 2. Relative to the predictive control, it replaces the binary outcomes
and their trace with a graded trace-and-score target under the same data and SFT schedule.
\item \emph{Continued assessment SFT} continues Stage 1 from the graded-SFT checkpoint for the GPU-hours of the GRPO
stage. Because those hours include rollout generation, it receives more updates than GRPO.
\item \emph{Calibration-only GRPO} runs the Stage 2 schedule with $\alpha=0$ in \cref{eq:grpo}.
\item \emph{Permuted targets} give each candidate another candidate's complete trace and score from the same scene,
keeping data, format, sequence lengths and compute fixed. \emph{Scores-only} and \emph{traces-only} shuffles break one
of the two alignments.
\item \emph{Longer policy training} spends AAPT's compute downstream instead, on the base-initialized policy.
\end{itemize}

\section{Results}
\label{sec:results}

\subsection{Is the assessment target meaningful?}
\label{sec:target}
We bin $1000$ held-out candidates by $\rgt$ and execute each in simulation with its own gripper width, so the test covers
the full grasp although $\rgt$ ignores width. A trial succeeds only with a $5$ to $10$\,cm lift, a stable hold and
force closure. Success rises monotonically from $2.9\%$ for $\rgt<0.25$ to
$92.8\%$ for $\rgt\geq0.80$ (\cref{fig:ranking}a). The reward also orders candidates. For a random successful and a
random failed candidate, the successful one has the higher reward $89\%$ of the time (AUC $0.89$). Restricting the
comparison to candidates from the same set, which is the comparison GRPO makes, gives $83.6\%$ (95\% CI $80.3$ to
$87.1$), so the reward orders local alternatives and not only candidates of different objects or scenes. Pooled agreement is
highest at the success boundary ($92\%$ between the two top bins). Both errors expected of a geometric proxy occur.
$29$ of $369$ candidates with $\rgt<0.5$ succeed, consistent with valid grasps missing from the annotations, and $21$
of $292$ with $\rgt\geq0.8$ fail (\supp{app:candidates}).

The base backbone ranks candidates barely above chance (\cref{fig:ranking}b). SFT learns a useful ordering, and GRPO
adds $5.4$, $8.0$ and $9.1$ points of pairwise accuracy from Test A to Test C. The tests differ in set size ($20$ to $75$) and also in the number and share of annotated
successes, so this trend shows that ranking refinement helps more on the larger sets but does not isolate set size. Depth improves global ordering, while the physical reasoning trace
improves discrimination among the best candidates (\cref{fig:ranking}c).

\subsection{What transfers?}
\label{sec:what}

\begin{table*}[t]
\centering
\footnotesize
\setlength{\tabcolsep}{4pt}
\begin{tabular}{@{}lcccccccc@{}}
\toprule
\textbf{Pre-training objective} & \makecell{\textbf{Action}\\\textbf{input}} & \makecell{\textbf{Failure}\\\textbf{examples}} &
\makecell{\textbf{Physical}\\\textbf{trace}} & \makecell{\textbf{Graded}\\\textbf{score}} & \makecell{\textbf{Ranking}\\\textbf{reward}} &
\makecell{\textbf{Targets}\\\textbf{aligned}} & \textbf{LIBERO-Plus} & \textbf{CALVIN} \\
\midrule
Base VLM, no pre-training & -- & -- & -- & -- & -- & -- & $87.0$ & $4.36$ \\
\midrule
\multicolumn{9}{@{}l}{\textit{Alternative objectives}}\\
Robo2VLM~\cite{robo2vlm} (embodied VQA) & \xmark & \xmark & \xmark & \xmark & \xmark & \cmark & $86.4$ & $4.38$ \\
VQASynth~\cite{vqasynth} (embodied VQA) & \xmark & \xmark & \xmark & \xmark & \xmark & \cmark & $87.3$ & $4.34$ \\
Grasp-pose prediction (successes only) & \xmark & \xmark & \xmark & \xmark & \xmark & \cmark & $85.9$ & $4.32$ \\
\midrule
\multicolumn{9}{@{}l}{\textit{Matched AAPT progression (same scenes, candidates, inputs)}}\\
Collision and stability prediction & \cmark & \cmark & \cmark & \xmark & \xmark & \cmark & $87.5$ & $4.42$ \\
Graded SFT (AAPT, Stage 1 only) & \cmark & \cmark & \cmark & \cmark & \xmark & \cmark & $88.1$ & $4.45$ \\
\quad + continued assessment SFT, GPU-hours of Stage 2 & \cmark & \cmark & \cmark & \cmark & \xmark & \cmark & $88.2$ & $4.46$ \\
\quad + calibration-only GRPO ($\alpha=0$) & \cmark & \cmark & \cmark & \cmark & \xmark & \cmark & $88.3$ & $4.48$ \\
\rowcolor{blktwo} AAPT (ranking + calibration GRPO) & \cmark & \cmark & \cmark & \cmark & \cmark & \cmark &
$\mathbf{88.8}$ & $\mathbf{4.54}$ \\
\midrule
\multicolumn{9}{@{}l}{\textit{Alignment control}}\\
AAPT, permuted targets & \cmark & \cmark & \cmark & \cmark & \cmark & \xmark & $82.6$ & $4.27$ \\
\midrule
\multicolumn{9}{@{}l}{\textit{Paired within-seed increments (mean $\pm$ SD over three seeds, positive on every seed)}}\\
\multicolumn{7}{@{}l}{Collision and stability $-$ Base} & $+0.5\pm0.2$ & $+0.06\pm0.02$ \\
\multicolumn{7}{@{}l}{Graded SFT $-$ Collision and stability} & $+0.7\pm0.2$ & $+0.04\pm0.01$ \\
\multicolumn{7}{@{}l}{AAPT $-$ Graded SFT} & $+0.7\pm0.2$ & $+0.09\pm0.02$ \\
\multicolumn{7}{@{}l}{AAPT $-$ Base} & $+1.8\pm0.2$ & $+0.18\pm0.03$ \\
\bottomrule
\end{tabular}
\caption{What transfers. All rows use Qwen3-VL-4B and the same ACoT-VLA recipe (mean over three seeds). LIBERO-Plus is
zero-shot success (\%), CALVIN is average sequence length. The matched progression keeps data and inputs fixed while
changing the principal quality supervision and then the ranking objective.
\supp{app:objective-audit} audits data, targets and budgets for every row, and \supp{app:paired} gives per-seed values.}
\label{tab:objective}
\end{table*}

\cref{tab:objective} organizes the matched objective progression, and its last block gives the paired within-seed
differences. Descriptive objectives and success-only grasp-pose prediction do not improve both benchmarks.

\begin{table}[t]
\centering
\footnotesize
\setlength{\tabcolsep}{3pt}
\begin{tabular}{@{}lccc@{}}
\toprule
\textbf{Seed 1} & \textbf{LIBERO} & \textbf{LIB.-Plus} & \textbf{CALVIN} \\
\midrule
Base, $60$K policy steps & $97.1$ & $86.4$ & $4.35$ \\
Base, $90$K steps (AAPT's compute) & $97.4$ & $86.7$ & $4.38$ \\
\midrule
Scores-only shuffle (own trace) & -- & $86.6$ & $4.31$ \\
Traces-only shuffle (own score) & -- & $85.9$ & $4.29$ \\
Full permutation & -- & $81.9$ & $4.25$ \\
\midrule
Candidate-only AAPT (no images) & -- & $86.2$ & $4.37$ \\
\midrule
\rowcolor{blktwo} AAPT, $60$K policy steps & $\mathbf{97.8}$ & $\mathbf{88.0}$ & $\mathbf{4.52}$ \\
\bottomrule
\end{tabular}
\caption{Compute and alignment controls on training seed 1, against the same-seed base and AAPT runs. Candidate-only AAPT is
trained without scene images (\cref{sec:see}).}
\label{tab:seed1}
\end{table}

\paragraph{Failures help, graded failures help more.} The collision-and-stability control sees exactly AAPT's
candidates and is supervised by physical outcomes measured in simulation, whereas AAPT never queries the simulator.
It already improves both benchmarks ($+0.5$ and $+0.06$, positive on every seed). Replacing its binary outcome
supervision with a graded trace-and-score target under the same SFT schedule adds $+0.7\pm0.2$ on LIBERO-Plus and
$+0.04\pm0.01$ on CALVIN, again
positive on every seed. Binary targets give a near-miss and a grasp that misses the object entirely the same label.
Graded targets tell them apart, consistent with the policy benefiting from their ordering.

\paragraph{The ranking reward, not more training.} Stage 2 adds a further $+0.7$ and $+0.09$. Spending the same
GPU-hours on continued SFT adds only $+0.1$ and $+0.01$, and GRPO with only the calibration term adds $+0.2$ and
$+0.03$. The ranking term therefore provides most of the Stage 2 gain, and on CALVIN its paired gain over
calibration-only GRPO is $+0.06\pm0.01$ on all three seeds. This matches \cref{fig:ranking}b, where ranking refinement
helps most on the larger candidate sets.

\paragraph{Alignment of both the trace and the score.} The permuted-target control keeps every input, output format
and the full budget, but gives each candidate another candidate's trace and score. It falls below the base, from
$87.0$ to $82.6$ and from $4.36$ to $4.27$, so the same data and compute are harmful once candidate and quality are
misaligned. In seed-1 diagnostics, corrupting either component (\cref{tab:seed1}) removes the full AAPT gain: an
aligned trace with a shuffled score stays near the base, while an aligned score with a shuffled trace falls below it.
Because each row retains one corrupted target, these tests establish sensitivity to joint alignment, not the
sufficiency of either component; \cref{tab:target} provides the clean score-only comparison.

\begin{table*}[t]
\centering
\footnotesize
\setlength{\tabcolsep}{3.5pt}
\begin{minipage}[t]{0.49\textwidth}
\centering
\begin{tabular}{@{}lccccc@{}}
\toprule
\multicolumn{6}{@{}l}{\textbf{(a) LIBERO success (\%)}}\\
\textbf{Method} & \textbf{Spatial} & \textbf{Object} & \textbf{Goal} & \textbf{Long} & \textbf{Avg} \\
\midrule
$\pi_0$~\cite{pi0} & 96.8 & 98.8 & 95.8 & 85.2 & 94.2 \\
$\pi_{0.5}$~\cite{pi05} & 98.8 & 98.2 & 98.0 & 92.4 & 96.9 \\
OpenVLA-OFT~\cite{openvla-oft} & 97.6 & 98.4 & 97.9 & 94.5 & 97.1 \\
VLA-Adapter~\cite{vla-adapter} & 97.8 & 99.2 & 97.2 & 95.0 & 97.3 \\
TurboVLA~\cite{turbovla} & 99.2 & 99.8 & 97.4 & 94.2 & 97.7 \\
MMaDA-VLA~\cite{mmada-vla} & 98.8 & 99.8 & 98.0 & 95.2 & 98.0 \\
ACoT-VLA~\cite{acot-vla} & 98.6 & 99.0 & 99.4 & 97.0 & 98.5 \\
\midrule
\rowcolor{blkone} VLM4VLA & 80.2 & 87.2 & 72.4 & 44.4 & 71.1 \\
\rowcolor{blkone} + AAPT & 84.0 & 90.2 & 77.6 & 60.8 & \textbf{78.2} \\
\rowcolor{blktwo} ACoT-VLA (ours) & 98.4 & 98.6 & 98.4 & 95.8 & 97.8 \\
\rowcolor{blktwo} + AAPT & 98.8 & 99.4 & 98.8 & 96.0 & \textbf{98.2} \\
\bottomrule
\end{tabular}
\end{minipage}\hfill
\begin{minipage}[t]{0.49\textwidth}
\centering
\begin{tabular}{@{}lcccccc@{}}
\toprule
\multicolumn{7}{@{}l}{\textbf{(b) CALVIN ABC$\rightarrow$D}}\\
\textbf{Method} & \textbf{1/5} & \textbf{2/5} & \textbf{3/5} & \textbf{4/5} & \textbf{5/5} & \textbf{Len.} \\
\midrule
$\pi_0$~\cite{pi0} & 96.8 & 85.0 & 76.7 & 68.1 & 59.9 & 3.92 \\
RoboVLMs~\cite{robovlms} & 98.0 & 93.6 & 85.2 & 77.9 & 70.4 & 4.25 \\
UniVLA~\cite{univla} & 98.9 & 94.8 & 89.0 & 82.6 & 75.1 & 4.41 \\
VLA-Adapter~\cite{vla-adapter} & 99.1 & 94.6 & 88.8 & 82.8 & 76.5 & 4.42 \\
DreamVLA~\cite{dreamvla} & 98.2 & 94.6 & 89.5 & 83.4 & 78.1 & 4.44 \\
MMaDA-VLA~\cite{mmada-vla} & 99.8 & 98.6 & 96.3 & 93.5 & 89.7 & 4.78 \\
\midrule
\rowcolor{blkone} VLM4VLA & 93.3 & 85.7 & 79.0 & 71.9 & 64.4 & 3.94 \\
\rowcolor{blkone} + AAPT & 94.4 & 88.1 & 83.2 & 77.4 & 70.9 & \textbf{4.14} \\
\rowcolor{blktwo} ACoT-VLA (ours) & 96.1 & 92.1 & 87.3 & 82.8 & 77.6 & 4.36 \\
\rowcolor{blktwo} + AAPT & 97.2 & 94.9 & 90.3 & 86.7 & 84.8 & \textbf{4.54} \\
\bottomrule
\end{tabular}
\end{minipage}
\caption{LIBERO and CALVIN ABC$\rightarrow$D. Shaded rows are our re-implementations on Qwen3-VL-4B (mean over three
seeds) and differ only in initialization. Published rows use their own backbones and data and give context only.
Full comparisons with seed SDs are in \supp{app:fulltables}.}
\label{tab:bench}
\end{table*}

\begin{table}[t]
\caption{SimplerEnv success (\%), visual-matching protocol, on the WidowX and Google Robot embodiments. Single
runs without seed uncertainty, so the comparison is descriptive.}
\label{tab:simpler}
\begin{center}
\footnotesize
\setlength{\tabcolsep}{4pt}
\begin{tabular}{lcccc}
\toprule
\textbf{Method} & \textbf{Carrot} & \textbf{Eggplant} & \textbf{Spoon} & \textbf{Cube} \\
\midrule
OpenVLA~\cite{openvla} &  4.2 &   0.0 &  0.0 & 12.5 \\
Octo~\cite{octo} &  8.3 &  43.1 & 12.5 &  0.0 \\
$\pi_0$~\cite{pi0}     & 62.5 & 100.0 & 54.2 & 25.0 \\
\midrule
\addlinespace[1pt]\multicolumn{5}{l}{\textbf{Ours}, on Qwen3-VL-4B}\\\addlinespace[1pt]
\rowcolor{blkone} VLM4VLA        & 54.2 &  95.8 & 75.0 &  0.0 \\
\rowcolor{blkone} AAPT + VLM4VLA & \textbf{58.4} & \textbf{100.0} & \textbf{79.2} & \textbf{4.2} \\
\rowcolor{blktwo} ACoT-VLA        & 62.5 & 100.0 & 83.3 & 33.3 \\
\rowcolor{blktwo} AAPT + ACoT-VLA & \textbf{70.8} & \textbf{100.0} & \textbf{87.5} & \textbf{45.8} \\
\bottomrule
\end{tabular}
\end{center}
\end{table}

\subsection{Is the gain explained by compute?}
\label{sec:compute}

AAPT pre-training (SFT and GRPO) costs about $2.3\times10^{19}$ FLOPs, roughly $0.6\times$ one downstream ACoT-VLA
run, or about $1\times$ including the one-time trace rendering (\supp{app:compute}). Several results show that
compute is not the source of the gain. The permuted-target control spends exactly AAPT's compute and hurts, and
continued assessment SFT for the GPU-hours of Stage 2 adds little. Finally, spending AAPT's compute on $30$K extra policy steps
($+50\%$, roughly the $38$K steps that AAPT's FLOPs would buy) moves the base by $+0.3$, $+0.3$ and $+0.03$ in the
single-seed diagnostic, while AAPT gains $+0.7$, $+1.6$ and $+0.17$ on that seed (\cref{tab:seed1}). A more useful exchange
rate is demonstrations. With half of the LIBERO demonstrations, AAPT reaches $87.2$ on LIBERO-Plus and already
exceeds the base trained on all of them ($87.0$), and AAPT pre-training is a one-time cost per backbone that is
reused across benchmarks and pipelines.

\subsection{Benchmark results}
\label{sec:bench}

AAPT improves both pipelines on every benchmark (\cref{tab:bench}, per-factor LIBERO-Plus results in \supp{app:fulltables}). The minimal VLM4VLA pipeline
gains $7.1$ points on LIBERO and $0.20$ on CALVIN. Its absolute performance reflects the deliberately minimal action
decoder. In ACoT-VLA, LIBERO
is near ceiling and the $+0.4$ gain mainly serves as a parity check. On the benchmarks with headroom, AAPT reaches
$88.8$ on zero-shot LIBERO-Plus and $4.54$ on CALVIN, behind only MMaDA-VLA, which uses $61$M steps of
cross-embodiment VLA pre-training. All nine paired differences over three seeds are positive (\supp{app:paired}). On
SimplerEnv, which uses the WidowX and Google Robot embodiments, AAPT improves every task not already at ceiling, most
on stack-cube with ACoT-VLA ($33.3$ to $45.8$, \cref{tab:simpler}).

\subsection{Where does the gain appear?}
\label{sec:where}

\begin{figure*}[t]
\centering
\includegraphics[width=0.82\textwidth]{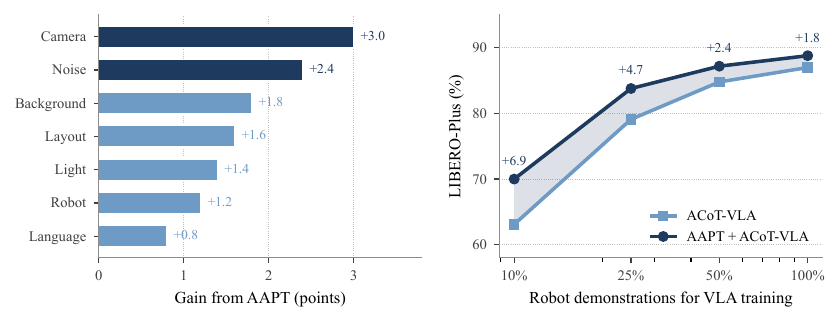}
\caption{AAPT gains with ACoT-VLA. \textbf{Left}, per-factor gain on zero-shot LIBERO-Plus. \textbf{Right},
LIBERO-Plus success as LIBERO demonstrations are withdrawn. The gap grows from $+1.8$ with all data to $+6.9$ at
$10\%$.}
\label{fig:gain}
\end{figure*}

If AAPT taught grasp-specific skills, gains would concentrate on grasping. They do not. With VLM4VLA, where the
backbone carries most of the policy, the gain is smallest on LIBERO-Object, which is pick-and-place ($+3.0$), and on
LIBERO-Spatial ($+3.8$). It is larger on LIBERO-Goal ($+5.2$), which includes opening a drawer, turning on a stove and
pushing a plate, and largest on LIBERO-Long ($+16.4$), whose multi-step tasks combine grasping with articulated objects
and other contact. On CALVIN, whose chains mix grasping with sliding, pushing and switching, the conditional per-step
gain grows toward the end of the chain and reaches $+4.1$ points at the fifth step with ACoT-VLA
(\supp{app:conditional}).

\paragraph{Robustness to viewpoint and sensor changes.} On zero-shot LIBERO-Plus all seven factors improve
(\cref{fig:gain}, left). Camera viewpoint gains most ($+3.0$), followed by sensor noise ($+2.4$), while language
rephrasing gains least ($+0.8$). This pattern is consistent with AAPT changing how the policy processes the visual
scene rather than instruction semantics, although it does not by itself identify which visual cues the policy uses.

\paragraph{Scarce demonstrations.} The advantage grows as demonstrations are withdrawn, from $+1.8$ with all data to
$+2.4$, $+4.7$ and $+6.9$ at $50\%$, $25\%$ and $10\%$ (\cref{fig:gain}, right). At $10\%$, AAPT reaches $70.0$,
roughly what the base reaches with $16$ to $17\%$ of the data. Intermediate checkpoints show that AAPT is ahead
throughout training ($+8.7$ at $20$K steps, $+6.4$ at $40$K, $+6.9$ at $60$K). Part of the advantage is faster
learning, since AAPT at $40$K already exceeds the final base, but the gap persists as both curves flatten, so AAPT
also reaches a higher level (\supp{app:lowdata}).

\subsection{Does the assessment use the scene?}
\label{sec:see}
Three inference-time controls test whether AAPT's ranking depends on the scene rather than on pose or
candidate-generation priors (\supp{app:audit}). With the candidate grasp alone (pose and width as text, no images), all-versus-all accuracy falls from $86.9$,
$80.1$ and $74.1$ on Tests A, B and C to $62.5$, $63.1$ and $59.8$, only a few points above the untrained backbone.
Pairing each pose with another scene's images lowers Test A accuracy to $57.2$, so a mismatched scene actively misleads
the model. Mirroring the images and the pose together keeps accuracy at $87.4$, $78.6$ and $69.8$. The
ranking therefore relies on the scene that matches the candidate. This grounding matters for transfer. AAPT trained
on the same targets without scene images leaves the policy at the base level on seed 1 ($86.2$ and $4.37$ against
$86.4$ and $4.35$), while full AAPT reaches $88.0$ and $4.52$ (\cref{tab:seed1}).

On $100$ unseen-scene candidates, AAPT generates traces from RGB, depth and the candidate grasp without the scaffold, mesh or
$\rgt$. An LLM judge compares each field with the scaffold, and we manually validate every judgment
(\supp{app:audit}). Object-dependent relations are correct in $82\%$ of cases, while $19\%$ of traces contain an
unsupported claim. Because objects are not held out, this audit establishes trace fidelity, not how the model obtains
the object information.

\begin{table}[t]
\centering
\footnotesize
\setlength{\tabcolsep}{3pt}
\begin{tabular}{@{}lccc@{}}
\toprule
\textbf{Field (vs.\ ground truth)} & \textbf{Tolerance} & \textbf{Match} & \textbf{Contradiction} \\
\midrule
\multicolumn{4}{@{}l}{\textit{Computable from the pose alone}}\\
Approach direction (camera frame) & exact & 87 & 12 \\
\multicolumn{4}{@{}l}{\textit{Requires the object's pose and geometry}}\\
Grasp type (vs.\ object up axis) & exact & 86 & 10 \\
Contact height & exact & 79 & 9 \\
Lateral offset from object axis & $\pm5$\,mm & 73 & 21 \\
Gripper fit & exact & 90 & 10 \\
Mean, object-dependent fields & & \textbf{82.0} & \textbf{12.5} \\
\bottomrule
\end{tabular}
\caption{Trace fidelity on $100$ held-out candidates (\%). The mean covers only the object-dependent fields.
Approach direction follows from the pose. The opening width is an input, so its
restatement is not audited here (\supp{app:audit}). Remaining cases omit the field.}
\label{tab:audit}
\end{table}

\begin{table}[t]
\centering
\footnotesize
\setlength{\tabcolsep}{3pt}
\begin{tabular}{@{}llcc@{}}
\toprule
\textbf{Input} & \textbf{Output target} & \textbf{Rank acc.} & \textbf{LIBERO-Plus} \\
\midrule
\multicolumn{2}{@{}l}{Base VLM, no AAPT} & -- & $87.0$ \\
RGB & score only & -- & $87.8$ \\
RGB + D & score only & $82.8$ & $87.2$ \\
RGB & trace $\rightarrow$ score & -- & $88.5$ \\
RGB + D & template $\rightarrow$ score & $85.7$ & $88.1$ \\
\rowcolor{blktwo} RGB + D & rewritten trace $\rightarrow$ score & $\mathbf{86.9}$ & $\mathbf{88.8}$ \\
\bottomrule
\end{tabular}
\caption{What in the target matters. Rank acc.\ is all-pairs accuracy on Test A; LIBERO-Plus is zero-shot success
with ACoT-VLA (three-seed mean). All AAPT rows also receive the candidate grasp; full metrics are in \supp{app:ranking}.}
\label{tab:target}
\end{table}

The template recovers most of the gain over score-only supervision ($+2.9$ ranking, $+0.9$ LIBERO-Plus), and rewriting
adds $+1.2$ and $+0.7$ (\cref{tab:target}). The template is computed from geometry alone and its renderer never sees
$\rgt$, so most of the trace's benefit does not depend on target information in its wording. The further gain from
rewriting may reflect better language, better selection of fields or cues from the reward the teacher sees.
Scaling to $1$M examples keeps improving both benchmarks with diminishing
returns (\supp{app:corpus}).

\section{Limitations and conclusion}
\label{sec:conclusion}

AAPT is limited to parallel-jaw grasps, object meshes and annotated successes; missing valid grasps can receive low
rewards, and the target does not grade candidates that differ only in gripper width. The assessment split holds out scenes but not objects, and simulation omits real perception and contact
stochasticity. Extensions to placement, trajectory-dependent actions, real robots and an action-pre-trained VLA such
as $\pi_{0.5}$ are important next tests.

Demonstration pipelines generate many near-misses and failures and then discard them. Learned through a graded,
candidate-aligned evaluative objective, these alternatives teach the backbone where the boundary lies between a
viable action and a nearby failure, and that makes it a better starting point for control.

\section*{LLM Usage Statement}
We used LLMs to assist with experimentation, data curation, auditing the evaluation sets
for leakage into pretraining data, figure plotting, identifying related work, and writing and framing the text. After
multiple rounds of revision, the authors edited and rewrote the final text themselves and checked every number, claim
and reference in the paper by hand before submission.

{
    \small
    \bibliographystyle{ieeenat_fullname}
    \bibliography{references}
}

\clearpage
\onecolumn
\raggedbottom
\appendix
\makeatletter\let\old@psect\psect\def\psect{\FloatBarrier\old@psect}\makeatother
\setcounter{figure}{0}
\setcounter{table}{0}
\setcounter{equation}{0}
\renewcommand{\thefigure}{S\arabic{figure}}
\renewcommand{\thetable}{S\arabic{table}}
\renewcommand{\theequation}{S\arabic{equation}}
\renewcommand{\theHfigure}{S\arabic{figure}}
\renewcommand{\theHtable}{S\arabic{table}}
\renewcommand{\theHequation}{S\arabic{equation}}

\begin{center}
{\Large\bfseries Appendix\par}
\end{center}
\vspace{0.5em}

\noindent Section, table and figure numbers without an ``S'' or a letter refer to the main paper.

\section{Notation}
\label{app:notation}

Table~\ref{tab:notation} collects the symbols used in Section~\ref{sec:method} and the appendices.

\begin{table}[!htbp]
\caption{Notation.}
\label{tab:notation}
\begin{center}
\small
\begin{tabular}{p{0.22\textwidth}p{0.70\textwidth}}
\toprule
\textbf{Symbol} & \textbf{Meaning} \\
\midrule
\multicolumn{2}{l}{\textit{Assessment task}}\\
$I_{\mathrm{RGB}}$, $I_{\mathrm{depth}}$ & Scene RGB image and optional depth image, VLM inputs \\
$g$, $g_i$ & Candidate grasp, a $6$-DoF pose and a gripper width, given to the VLM as text (position, quaternion and width) \\
$x$, $y$ & VLM input $(I_{\mathrm{RGB}},I_{\mathrm{depth}},g)$ and output $(z_{\mathrm{phys}},\texttt{<score>},s)$ \\
$z_{\mathrm{phys}}$ & Physical reasoning trace; an output target constructed offline, never an input \\
$s$, $s_i$ & Scalar score generated after \texttt{<score>}; $s_i$ is the score of one rollout for $g_i$ \\
\midrule
\multicolumn{2}{l}{\textit{Target reward (Equation~\ref{eq:reward})}}\\
$\mathbf{T}=(\mathbf{R},\mathbf{t})$ & Grasp pose as rotation and translation \\
$d_{\tau}$ & Scaled $\mathrm{SE}(3)$ distance between two poses, symmetric under the gripper's $180^\circ$ flip \\
$\tau_R$, $\tau_t$ & Rotational and translational reference scales, $0.02$\,rad and $0.02$\,m \\
$d_{\min}$ & Distance from a candidate to the nearest annotated success in the scene \\
$\tilde d$ & Median distance over the scene's failures \\
$r^{\mathrm{gt}}$ & Graded target reward in $[0,1]$; seen only by the offline teacher and the training losses \\
\midrule
\multicolumn{2}{l}{\textit{Ranking refinement (GRPO)}}\\
$\bar s_i$, $\sigma_i$ & Mean and standard deviation of the scores in the five-rollout group for $g_i$ \\
$p_{\mathrm{gt}}$, $p_{\mathrm{pred}}$ & Target and predicted probabilities that $g_i$ ranks above $g_j$ \\
$\sigma(\cdot)$, $\Phi(\cdot)$ & Logistic function and standard normal CDF \\
$\kappa$, $\gamma$, $\lambda$ & Preference sharpness ($8$), variance regularizer ($0.01$) and variance floor ($0.25$) \\
$\sigma_{\mathrm{comb}}$ & Combined standard deviation used in $p_{\mathrm{pred}}$ \\
$K$ & Number of nearest-rank comparison candidates ($4$) \\
$r_{\mathrm{pair}}$, $R_{\mathrm{pair}}$ & Pairwise agreement reward and its average over the $K$ comparisons \\
$R$, $\alpha$ & Rollout reward and its ranking weight ($0.6$) \\
\midrule
\multicolumn{2}{l}{\textit{Downstream policies}}\\
$H$, $H^{\mathrm{ref}}$ & Action horizon and ACoT-VLA reference-trajectory horizon \\
\bottomrule
\end{tabular}
\end{center}
\end{table}

\section{Candidate set generation}
\label{app:candidates}

Each object instance yields a candidate set containing an annotated success, controlled perturbations of that
success and clear failures. The construction provides both difficult comparisons near a known success and
unambiguous low-quality candidates.

\paragraph{Successful poses.}
We sample one annotated successful grasp as the seed for each candidate set and assign the unperturbed seed
$r^{\mathrm{gt}} = 1$. HouseCat6D supplies binary success labels directly. GraspNet-1Billion supplies continuous
force-closure scores, which we threshold at $0.8$ to define the annotated success set. Scores below the threshold
are treated as failures. An object can have thousands of valid grasps, so we do not use every success as a
perturbation seed. Seeds are sampled randomly subject to diversity over $\mathrm{SE}(3)$, which avoids dense
clusters of near-duplicate poses and covers distinct approach and contact configurations. All annotated successes
remain available when computing the reward, even though only one is selected as the seed of a given set.

\paragraph{Near-successful perturbations.}
We generate four $\mathrm{SE}(3)$-perturbed variants of that grasp at increasing noise levels
$\sigma_R \in \{0.02, 0.05, 0.08, 0.12\}$ radians and
$\sigma_t \in \{0.02, 0.05, 0.08, 0.12\}$ meters. Rotation uses zero-mean isotropic Gaussian axis-angle noise and
translation uses an independent zero-mean isotropic Gaussian in $\mathbb{R}^3$; the four rungs pair the listed
rotation and translation standard deviations. All four perturbations keep the seed's gripper width, so candidates
never differ only in width.
These are the candidates that straddle the success boundary, and their rewards decay smoothly with
$\mathrm{SE}(3)$ distance to the nearest success.

\paragraph{Failure poses.}
The remaining candidates are sampled from the dataset's annotated failures. If an object does not provide enough
failure poses, we fill the remaining slots with uniformly sampled poses. This fallback ensures that every set also
contains candidates far from any annotated success.

\paragraph{Training-time composition.}
Every training candidate set contains $N=20$ poses. It includes one unperturbed seed, one perturbation from each of
the four noise rungs and $15$ failure poses from the dataset or the uniform fallback. Corpus sizes in the paper
count individual candidate examples rather than 20-candidate sets. Across the corpus, $40\%$ of examples come from
HouseCat6D and $60\%$ from GraspNet-1Billion.

\paragraph{Reward assignment.}
Every candidate is scored by Equation~\ref{eq:reward} using its minimum $\mathrm{SE}(3)$ distance to any annotated
success for that object in the scene. The rotational term is the smaller of the geodesic angles to the success and to
the success rotated by $180^\circ$ about the gripper's approach axis, because a parallel-jaw grasp is unchanged by that
rotation. The gripper width does not enter the distance. The rotational and translational reference scales first make the two distance
terms dimensionless. Both scales equal the standard deviation of the smallest perturbation rung, so a
reference-scale rotational error and a reference-scale translational error contribute equally. Dividing by the
median failure distance then adapts the reward range to the scene. A candidate at the median failure distance has
$r^{\mathrm{gt}}=0.5$, while more distant failures approach zero.

Taking the minimum over all annotated successes accommodates multiple valid grasps, including symmetry-equivalent
ones and successes not selected as perturbation seeds. It cannot account for missing annotations, so an
unrecorded valid grasp may receive an artificially low reward. The target is therefore a relative geometric quality
score, not a calibrated success probability. Annotated successes and nearby perturbations occupy the high-reward
end, while distant failures anchor the low-reward end. This mixture yields easy comparisons against gross failures
and difficult comparisons among geometrically similar poses near the success boundary. The GRPO stage
therefore compares each candidate only with its nearest $K$ neighbors in ground-truth rank.

\paragraph{Evaluation splits.}
We hold out $200$ scenes for assessment evaluation and construct three candidate-set sizes per object instance.
Test A contains $20$ candidates with $5$ successes and $15$ failures. Test B contains $50$ with $20$ successes and
$30$ failures. Test C contains $75$ with $25$ successes and $50$ failures. These successes are annotated dataset
grasps at $r^{\mathrm{gt}}=1$, not synthesized perturbations. The evaluation sets therefore draw several annotated
successes per object instance, whereas each training set is seeded from a single one, and no perturbed pose is ever
counted as a success at test time. The larger sets contain more plausible
alternatives and therefore more difficult pairwise decisions. The split is scene-disjoint but not object-instance-
disjoint or category-disjoint. These diagnostics measure generalization to unseen scenes, not to unseen object
identities.

\paragraph{Physical validation of the graded reward.}
We execute $1000$ synthetic candidates in simulation and measure empirical grasp success within fixed
$r^{\mathrm{gt}}$ bins. These are observed simulator outcomes, not predictions from AAPT or any control. The
simulator executes each candidate with its own gripper width. The
binary success label requires the object to be lifted $5$--$10$\,cm, remain stably held over the fixed evaluation
window and pass force-closure verification. The
candidates are drawn from a set held out entirely from the pre-training corpus, so no
executed candidate contributes to any AAPT, control or baseline training run. The validation therefore tests the
target construction itself and cannot leak simulator outcomes into a trained model.
Annotated seed successes at $r^{\mathrm{gt}}=1$ are excluded. Ordinary candidate generation
does not populate every interval equally, so we resample the generation process until all bins have adequate
coverage. This stratification changes the reward distribution and makes a pooled success rate uninterpretable.
Table~\ref{tab:reward-validity} therefore reports the number of executions and conditional success rate within each
bin. Every candidate uses the same joint 6-DoF perturbation process as pre-training. Translation and rotation are
not varied separately, and the analysis is not stratified by object symmetry.

\begin{table}[!htbp]
\caption{Physical validation of the geometric reward over $1000$ synthetic simulator executions. Success increases
monotonically with $r^{\mathrm{gt}}$. Sampling was stratified to cover all bins; consequently, the per-bin rates, not
their pooled average, are the relevant quantities. Intervals are $95\%$ Wilson score intervals on the per-bin rate.}
\label{tab:reward-validity}
\begin{center}
\small
\begin{tabular}{lrrrc}
\toprule
\textbf{Reward bin} & \textbf{Candidates} & \textbf{Successes} & \textbf{Empirical success (\%)} & \textbf{95\% CI} \\
\midrule
$0 < r^{\mathrm{gt}} < 0.25$       & 170 & 5   & 2.94  & [1.3,\ 6.7] \\
$0.25 \leq r^{\mathrm{gt}} < 0.50$ & 199 & 24  & 12.06 & [8.2,\ 17.3] \\
$0.50 \leq r^{\mathrm{gt}} < 0.67$ & 113 & 27  & 23.89 & [17.0,\ 32.5] \\
$0.67 \leq r^{\mathrm{gt}} < 0.80$ & 226 & 116 & 51.33 & [44.8,\ 57.8] \\
$0.80 \leq r^{\mathrm{gt}} < 1.00$ & 292 & 271 & 92.81 & [89.3,\ 95.2] \\
\midrule
\textbf{Total} & \textbf{1000} & \textbf{443} & -- & -- \\
\bottomrule
\end{tabular}
\end{center}
\end{table}

\paragraph{Ordering agreement with simulator outcomes.}
The same executions also measure how well $r^{\mathrm{gt}}$ orders candidates. Over all pairs of one successful and
one failed candidate, the successful one has the higher reward in $89\%$ of pairs (AUC $0.89$). Restricting to pairs
drawn from adjacent bins of Table~\ref{tab:reward-validity}, the higher-reward candidate is the successful one in
$82\%$, $70\%$, $77\%$ and $92\%$ of pairs across the four boundaries, from lowest to highest. Agreement is weakest
between $[0.25,0.50)$ and $[0.50,0.67)$, the adjacent pair whose intervals overlap, and strongest at the success
boundary between $[0.67,0.80)$ and $\geq0.80$. Both error types expected of a geometric proxy occur: $29$ of the
$369$ candidates with $r^{\mathrm{gt}}<0.5$ succeed, consistent with valid grasps missing from the annotations, and
$21$ of the $292$ candidates with $r^{\mathrm{gt}}\geq0.8$ fail. These figures pool objects and treat candidates in the
same bin as ties. Pooling can benefit from differences between objects and scenes, whereas AAPT and its GRPO stage
only compare candidates within a set, which share one object, one scene and one success set. Restricted to pairs of a
successful and a failed candidate from the same set, the successful candidate has the higher reward in $83.6\%$ of
pairs (95\% bootstrap CI over sets, $80.3$ to $87.1$). Most of the pooled agreement therefore holds among local
alternatives.

\section{Physical reasoning trace generation}
\label{app:traces}

The physical reasoning trace target is constructed offline in two steps. A deterministic generator first converts
available ground-truth geometry into a structured scaffold. Qwen3-VL-8B then rewrites that scaffold as concise
natural language. The two datasets expose different geometric metadata, so their scaffold generators differ.

\subsection{HouseCat6D}
\label{app:traces-housecat}

HouseCat6D provides binary success labels and a categorical object name such as \emph{bottle}, \emph{cup} or
\emph{teapot}. Given the object's camera-frame pose $\mathbf{T}_{\mathrm{obj}}$, its scale $\mathbf{s} = (w, h, d)$,
the candidate grasp transform $\mathbf{T}_{\mathrm{grasp}}$ and optionally the depth image, the generator computes

\begin{enumerate}
  \item \textbf{Grasp-center direction}, by classifying the unit vector from the object centroid to the grasp center
    as above, below, left, right, front, behind or angled.
  \item \textbf{Grasp type}, from the dot product of the gripper approach axis
    $\mathbf{R}_{\mathrm{grasp}}\hat{\mathbf{z}}$ with the object's up axis, giving top-down, bottom-up, side or
    angled.
  \item \textbf{Shape descriptor}, from the aspect ratio $\max(\mathbf{s})/\min(\mathbf{s})$, giving elongated,
    compact or intermediate-aspect-ratio.
  \item \textbf{A category-specific part vocabulary}, where the object name selects a target part such as the neck,
    the rim, the handle or the toe box, together with a grasp-strategy phrase consistent with the grasp-center
    direction.
  \item \textbf{Relative height and lateral offset}, namely where along the object's vertical axis the grasp lands
    and how far it sits off the central axis.
  \item \textbf{Depth analysis}, when the depth image is available, reporting the object's distance from the camera,
    the clearance between approach and contact, and a finger-spacing hint of narrow, wide or optimal.
\end{enumerate}

\subsection{GraspNet-1Billion}
\label{app:traces-graspnet}

GraspNet-1Billion has no complete category vocabulary because only $34$ of its $88$ objects carry semantic YCB
names. Every object does provide a mesh. Once per object identifier, we compute its axis-aligned extents, PCA
principal axes, oriented extents and a cross-section similarity score
$\mathrm{sym} = \min(e_2, e_3)/\max(e_2, e_3)$. Values near one indicate similar extents along the two minor
principal axes and serve as a limited proxy for axial symmetry around the longest axis. They are not a complete
test of object symmetry. We also assign one of six shape tags, namely elongated-cylindrical, elongated, flat,
roughly-spherical, compact or intermediate-aspect-ratio. For each candidate, the generator reports

\begin{enumerate}
  \item \textbf{The object's pose in the camera frame}, classified as upright, upside-down, tilted on its side or
    tilted at an angle by projecting the model-up axis into the camera frame.
  \item \textbf{The camera-frame approach direction}, classified as top-down, bottom-up or from the left, right,
    front or rear using the camera-frame gripper $+\mathbf{z}$ axis.
  \item \textbf{The object-frame approach alignment}, obtained by re-expressing the approach direction in the object
    frame and testing it against the longest, intermediate and shortest principal axes.
  \item \textbf{Grasp type relative to the object's up axis}, giving top-down, bottom-up, side or angled.
  \item \textbf{Contact location}, by transforming the grasp center into the object frame and reporting relative
    height along with the lateral offset from the vertical axis in millimeters.
  \item \textbf{Gripper fit}, by comparing the opening width against the object's thinnest axis to give too narrow,
    well matched, reasonable or too wide, with the depth the gripper is translated into the object reported in
    millimeters.
\end{enumerate}

Where the object identifier appears in the known YCB mapping we prepend the name. Otherwise the object is referred to
only by its shape and dimensions, which avoids confabulating names for the $54$ objects that have none.

\begin{figure}[!htbp]
\begin{tcolorbox}[colback=gray!5,colframe=gray!40,title=Procedural scaffold output for two candidates on the same object]
\footnotesize
\textbf{Object.} Peach, roughly-spherical, $6.2 \times 6.3 \times 5.9$\,cm, symmetry score $1.00$.\\[4pt]
\textbf{Highest-reward candidate}, grasp width $6.3$\,cm. \emph{The target is a peach, classified as roughly-spherical
(dimensions 6.2 x 6.3 x 5.9 cm, oriented-extent 6.4/5.9/5.9 cm, symmetry score 1.00). In this view the object sits
tilted at an angle at 0.48 m depth from the camera. The candidate gripper approaches top-down, aligned with the
object's longest axis. This is an angled grasp, placed near the top of the object, 55 mm off-center laterally. The
gripper opening (6.3 cm) is well matched to the object's thinnest axis (5.9 cm); the gripper is translated 30 mm into
the object along the approach axis.}\\[4pt]
\textbf{Lowest-reward candidate}, same instance, grasp width $9.4$\,cm. \emph{\ldots The candidate gripper approaches from
below or rear, aligned with the object's intermediate axis. This is a side grasp, perpendicular to the object's up
axis, placed near the top of the object, slightly off-center laterally (28 mm). The gripper opening (9.4 cm) is
reasonable for the object's thinnest axis (5.9 cm); the gripper is translated 10 mm into the object along the approach
axis.}
\end{tcolorbox}
\caption{What the procedural generator emits for two candidates on the same object instance. This scaffold is used
offline to construct the model's output target; it is never included in the model prompt. Sec.~\ref{app:traces-render}
rewrites it into the target reasoning trace. The entries share the same object description, while their
candidate-specific approach, contact and gripper-fit fields differ. Object identity and appearance alone therefore
cannot determine their ordering.}
\label{fig:scaffoldex}
\end{figure}

\subsection{Rendering the scaffold as text}
\label{app:traces-render}

The deterministic templates above produce structured language grounded in geometry, but we do not use that text
directly as the training target. We provide the scaffold and its supporting metadata to Qwen3-VL-8B for one offline
generation per example. The rewrite converts the fixed fields into concise natural language while preserving their
physical content. The resulting text is the VLM's autoregressive \emph{output target}, not an input feature.
Figure~\ref{fig:traceprompt} gives the prompt that turns a scaffold such as the one in Figure~\ref{fig:scaffoldex}
into that target trace.

\paragraph{What the trace can encode.} The scaffold is computed from geometry alone. Its qualitative descriptors,
such as gripper fit (too narrow, well matched, reasonable or too wide), come from fixed thresholds on the opening
relative to the object's thinnest axis and do not depend on $r^{\mathrm{gt}}$. The rewriting teacher, in contrast,
sees $r^{\mathrm{gt}}$ and is asked to describe evidence consistent with it. It never states the score, a tier or a
verdict, but it can still encode target information through which fields it emphasizes and how it words them. The
procedural-template variant of Table~\ref{tab:target} removes this channel, because its text is generated without
$r^{\mathrm{gt}}$, and it already recovers most of the trace's benefit (Table~\ref{tab:trace-rendering}). We
therefore read the trace as a distilled description of privileged geometry rather than as a faithful reasoning
process.

\begin{figure}[!htbp]
\begin{tcolorbox}[colback=gray!5,colframe=gray!40,title=Trace-rendering prompt for assessment reasoning]
\footnotesize
\textbf{System.} You are an expert in robotic grasping. You reason about grasp candidates based on object geometry,
gripper pose, and contact quality.\\[4pt]
\textbf{User.}\\
\textbf{Image 1 (RGB).} \texttt{[scene RGB observation]}\\
\textbf{Image 2 (Depth).} \texttt{[scene depth image]}\\
\textbf{Object description.} \texttt{[category, dimensions, principal axes from CAD / PCA]}\\
\textbf{Candidate metadata.} \texttt{[position, quaternion, gripper width]}\\
\textbf{Target reward $r^{\mathrm{gt}} \in [0,1]$.} \texttt{[ground-truth quality score]}\\
\textbf{Procedural scaffold.} \texttt{[structured analysis from
Secs.~\ref{app:traces-housecat}/\ref{app:traces-graspnet} covering approach direction, grasp type, contact location,
gripper fit]}\\[4pt]
\textbf{Produce a concise natural-language reasoning trace that implicitly justifies the target reward without
stating the score itself.} First establish the physical evidence by covering:\\
(i)~the object and its pose in the scene,\\
(ii)~the candidate grasp in terms of approach direction and grasp type,\\
(iii)~the contact location and gripper-object fit.\\[4pt]
Stay faithful to the procedural scaffold. State only physical evidence. Never give the score, a quality tier or any
qualitative rating of the grasp.
\end{tcolorbox}
\caption{Offline trace-generation prompt used to construct assessment-reasoning targets. Qwen3-VL-8B receives
privileged metadata, the scaffold and target reward, and writes only the physical evidence that justifies the
reward. The trace contains no score, tier or verdict, so the numerical score that follows it is the only quality
statement in the target.
The trained VLM receives the scene and the candidate grasp (pose and width), but not the object description, scaffold or
reward; it must generate the trace before emitting its score.}
\label{fig:traceprompt}
\end{figure}

\paragraph{Collision-and-stability control.}
The same-data predictive control of Table~\ref{tab:objective} does not use the prompt of
Figure~\ref{fig:traceprompt}. Its trace generator is never exposed to $r^{\mathrm{gt}}$. Its procedural scaffold contains the candidate's geometric fields and the two binary simulator outcomes,
which are its only quality information, and Figure~\ref{fig:traceprompt-control} gives its rendering prompt. Like AAPT's,
its traces state only the physical evidence.

\begin{figure}[!htbp]
\begin{tcolorbox}[colback=gray!5,colframe=gray!40,title=Trace-rendering prompt for the collision-and-stability control]
\footnotesize
\textbf{System.} You are an expert in robotic grasping. You reason about grasp candidates based on object geometry,
gripper pose, and contact quality.\\[4pt]
\textbf{User.}\\
\textbf{Image 1 (RGB).} \texttt{[scene RGB observation]}\\
\textbf{Image 2 (Depth).} \texttt{[scene depth image]}\\
\textbf{Object description.} \texttt{[category, dimensions, principal axes from CAD / PCA]}\\
\textbf{Candidate metadata.} \texttt{[position, quaternion, gripper width]}\\
\textbf{Simulator outcomes.} \texttt{[environment collision: yes / no; stable grasp: yes / no]}\\
\textbf{Procedural scaffold.} \texttt{[structured analysis from
Secs.~\ref{app:traces-housecat}/\ref{app:traces-graspnet} covering approach direction, grasp type, contact location,
gripper fit, simulator outcomes]}\\[4pt]
\textbf{Produce a concise natural-language reasoning trace that implicitly justifies the simulator outcomes without
stating them.} Establish the physical evidence by covering:\\
(i)~the object and its pose in the scene,\\
(ii)~the candidate grasp in terms of approach direction and grasp type,\\
(iii)~the contact location and gripper-object fit.\\[4pt]
Stay faithful to the procedural scaffold.
\end{tcolorbox}
\caption{Offline trace-generation prompt for the collision-and-stability control. Unlike
Figure~\ref{fig:traceprompt}, the teacher is never exposed to $r^{\mathrm{gt}}$. Its scaffold contains the same geometric fields as AAPT's, and the two binary simulator outcomes are its only
quality information. As for AAPT, the trace covers only the physical evidence (i)--(iii). The VLM's
prompt is identical to AAPT's (scene and candidate grasp only); its target is the trace followed by the two
binary indicators,
\texttt{<collision>}~$c$ \texttt{<stable>}~$b$ with $c,b\in\{\texttt{yes},\texttt{no}\}$, in place of
\texttt{<score>} and $r^{\mathrm{gt}}$.}
\label{fig:traceprompt-control}
\end{figure}

\paragraph{Exact autoregressive training sequence.}
The offline procedure above produces a fixed target trace $z_{\mathrm{phys}}$ for each candidate, but that trace is
never concatenated to the model prompt. In SFT, the model input is
\[
x = [I_{\mathrm{RGB}},\, I_{\mathrm{depth}},\, \text{candidate position},\, \text{candidate quaternion},\, \text{gripper width}],
\]
and the next-token target is
\[
y = [z_{\mathrm{phys}},\, \texttt{<score>},\, r^{\mathrm{gt}}].
\]
During next-token training, earlier tokens of $z_{\mathrm{phys}}$ appear only as preceding target tokens in the
usual autoregressive sequence. They are never placed in the prompt. During GRPO and assessment evaluation, the
model receives only $x$, generates $\hat z_{\mathrm{phys}}$ itself, and then emits
\texttt{<score>} and $\hat s$. The ranking metrics and GRPO reward use $\hat s$. During downstream VLA training,
neither $x$ nor this output format is retained. Only the learned backbone parameters initialize the policy.

\section{Optimization and score parsing}
\label{app:optimization}

For SFT, we update the language model at full rank while freezing the vision encoder. Training runs for one epoch
with learning rate $5\mathrm{e}{-}6$, batch size $32$, cosine decay and $10\%$ warmup. The prompt requests a decimal
score in $[0,1]$ after \texttt{<score>}. The parser reads the first numeric value after that sentinel. It does not
clip the parsed value. The output instruction is the only mechanism requesting the $[0,1]$ range. If a completion
contains no parseable value after the sentinel, it is discarded and regenerated with the same decoding settings.

The GRPO stage runs for $500$ steps at learning rate $3\mathrm{e}{-}6$ with five rollouts per candidate,
$\kappa=8$, $\gamma=0.01$, $\lambda=0.25$, $K=4$ and $\alpha=0.6$. We use a KL penalty of
$5\mathrm{e}{-}3$, weight decay $1\mathrm{e}{-2}$ and gradient clipping at $1.0$. Each GRPO group contains five
complete sequences sampled from the same scene-and-candidate prompt, and its rewards are normalized within that group.
When a rollout for candidate $g_i$ is compared with candidate $g_j$, $\bar{s}_j$ and $\sigma_j$ are computed from
the five separately generated completions for $g_j$. The uncertainty term $\sigma_i$ is computed from all five
completions for $g_i$, including the current rollout score $s_i$. Thus, the current rollout contributes to
$\sigma_i$ but never to the comparison mean $\bar{s}_j$. We transfer the final checkpoint after the fixed SFT and
GRPO budgets and do not select it using downstream policy performance.

\section{Downstream pipeline details}
\label{app:pipelines}

\begin{figure}[!htbp]
\centering
\includegraphics[width=\textwidth]{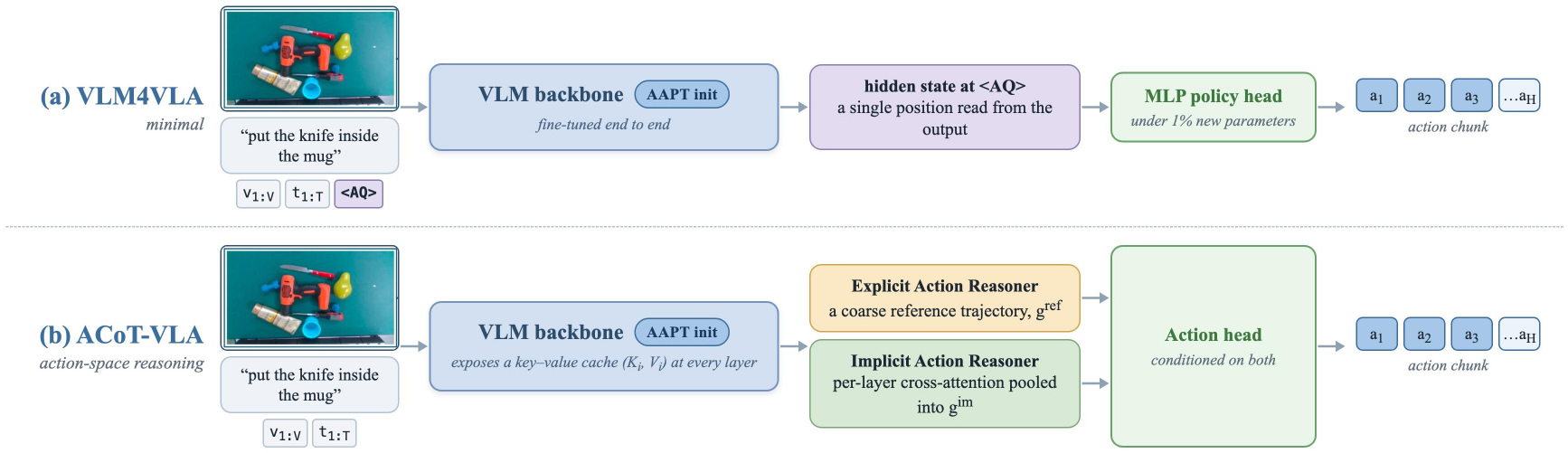}
\caption{The two downstream pipelines, both initialized from the same backbone.
\textbf{(a)} VLM4VLA~\cite{vlm4vla} appends one learnable action query token and decodes its hidden state through an
MLP. \textbf{(b)} ACoT-VLA~\cite{acot-vla} attaches two
reasoners to the key-value cache, one proposing a coarse reference trajectory and one pooling per-layer
cross-attention into a latent action prior, keeping the backbone representation directly involved even near ceiling.}
\label{fig:pipelines}
\end{figure}

Both pipelines take the current RGB observation and language instruction, with no proprioceptive or privileged
state. Each policy is initialized from the relevant pre-trained checkpoint and then trained on its downstream
benchmark.

\paragraph{VLM4VLA.}
A single learnable action-query token is appended to the token sequence. A two-layer MLP decodes the hidden state at
that position into the next $H$ end-effector actions, each represented by position, orientation and gripper state.
Training updates the complete model, including the backbone. Position and orientation use Huber regression, while
gripper state uses binary cross-entropy. The MLP produces the action chunk directly without a diffusion or
flow-matching decoder.

\paragraph{ACoT-VLA.}
The Explicit Action Reasoner is a compact Transformer with one reasoner layer for each backbone layer. Reasoner
layer $i$ cross-attends to the key-value cache of backbone layer $i$ and denoises a reference action sequence of
horizon $H^{\mathrm{ref}}$ into a coarse trajectory. The reference target is a temporally downsampled copy of the
demonstrated action chunk, so it requires no separately annotated reasoning labels. During training, the action head
is conditioned on explicit guidance computed from the ground-truth reference trajectory. At inference, the
reasoner instead conditions on its own predicted trajectory, following the stabilization protocol of the original
method.

The Implicit Action Reasoner attaches a learnable query to every backbone layer, pools the resulting cross-attention
output and projects it through an MLP. The projected outputs are aggregated into one guidance vector. This vector
has no separate supervision and is learned only through the downstream action loss. The explicit trajectory and
implicit vector jointly condition the action head.

We make two architectural adaptations for Qwen3-VL-4B. The published $\pi_{0.5}$-derived configuration pairs SigLIP
with an 18-layer Gemma-2B language model at $224\times224$ resolution. We instead use Qwen3-VL-4B at its native
dynamic resolution. The explicit reasoner therefore follows Qwen's layer count, and its cross-attention projections
match Qwen's grouped-query key-value layout.

\section{Downstream training and evaluation protocols}
\label{app:downstream-protocols}

\paragraph{Replication.}
All experiments use Qwen3-VL-4B unless stated otherwise, and AAPT uses the $500$K-sample corpus outside the
corpus-size sweep. Every baseline and AAPT comparison shares the downstream seed and data order. We use the same
three seeds for both pipelines and treat the training seed, not rollout episodes or perturbation factors, as the
replication unit. Since three seeds do not support strong inferential claims, we report paired differences
descriptively without a formal significance test. Sec.~\ref{app:paired} gives the exact matched values.

\paragraph{LIBERO and LIBERO-Plus.}
LIBERO~\citep{libero} contains Spatial, Object, Goal and Long suites, each with ten tasks and fifty human
demonstrations per task. We train one policy per suite and report mean success over fifty rollouts per task.
LIBERO-Plus~\citep{libero-plus} re-renders these tasks under seven held-out factors covering camera pose, robot
configuration, language phrasing, lighting, background, sensor noise and object layout. It contains $10{,}030$
episodes. We evaluate the LIBERO checkpoints without additional fine-tuning.

\paragraph{CALVIN.}
CALVIN ABC$\rightarrow$D~\citep{calvin} trains on environments A, B and C and evaluates in held-out environment D.
The protocol contains $1000$ chains of five language instructions. We report the rate of completing at least each
chain position and the mean number completed.

\paragraph{VLM4VLA training.}
The action horizon is $H=4$ on LIBERO and SimplerEnv and $H=10$ on CALVIN. On LIBERO and SimplerEnv, we fine-tune
all parameters with AdamW for $50$K steps using learning rate $5\mathrm{e}{-}5$ and batch size $512$. CALVIN uses
$60$K steps, learning rate $2\mathrm{e}{-}5$ and batch size $128$.

\paragraph{ACoT-VLA training.}
The reference horizon is $H^{\mathrm{ref}}=15$ and the action horizon is $H=10$. We train with AdamW for $60$K
steps using peak learning rate $5\mathrm{e}{-}5$, a $10$K-step warmup, cosine decay and batch size $128$.
Checkpoint selection follows the original ACoT-VLA protocol~\citep{acot-vla}.

\paragraph{SimplerEnv.}
\label{app:simpler}
Table~\ref{tab:simpler} in the main paper reports the SimplerEnv visual-matching protocol~\citep{simplerenv}. Its
WidowX and Google Robot embodiments differ from the Franka arm used in LIBERO and CALVIN. We report each task
separately because their operating points differ substantially. Put eggplant in basket is saturated for our
policies, while stack cube is near zero for VLM4VLA but improves with ACoT-VLA, and a four-task mean would obscure
both effects.

\section{Assessment task evaluation}
\label{app:ranking}

All results in this section use the $500$K-sample pre-training corpus and the $200$ held-out scenes defined in
Sec.~\ref{app:candidates}. We report two pairwise metrics. \emph{All-versus-all accuracy} is the fraction of valid
candidate pairs whose predicted ordering agrees with the ground-truth ordering. \emph{Top-$M$ accuracy} uses only
pairs containing at least one of the $M=5$ highest-reward candidates, emphasizing decisions that involve poses near
the top of the set. Chance accuracy is $50.0\%$ for either pairwise metric at every candidate-set size.

Exact ties do not contribute to pairwise metrics. We exclude a
pair if its ground-truth rewards are equal or its predicted scalar scores are equal.

\begin{figure}[!htbp]
\centering
\includegraphics[width=0.72\textwidth]{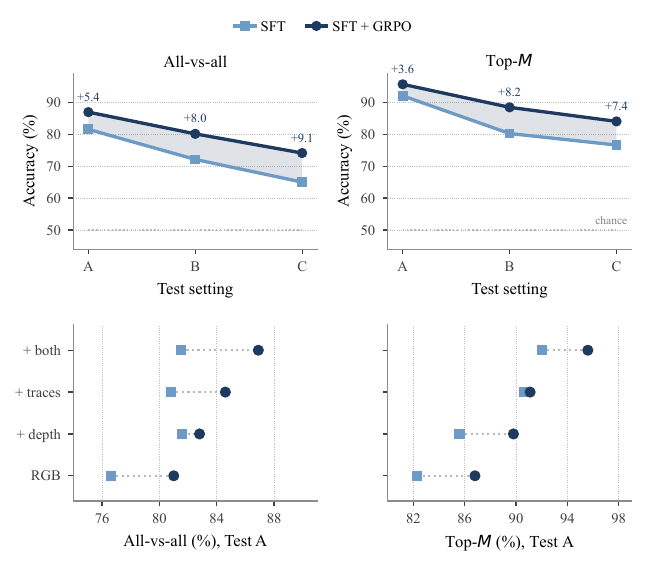}
\caption{Assessment accuracy on $200$ held-out scenes after training on $500$K samples, on both pairwise metrics. Top row, the three test settings, which
present $20$, $50$ and $75$ candidates per object. Bottom row, the component ablation on Test A, where PR denotes
supervision to generate a procedural physical-reasoning trace before the score. The trace is an output target, not an
input. Columns share a metric. Dotted lines in the top row mark chance ($50\%$).}
\label{fig:rankingfull}
\end{figure}

Figure~\ref{fig:rankingfull} expands the two ranking panels of Figure~\ref{fig:ranking} to both metrics, which shows
two things the aggregate hides.

First, the metrics respond differently across the tests. GRPO improves all-versus-all accuracy by $5.4$, $8.0$ and
$9.1$ points on Tests A, B and C, and Top-$M$ accuracy by $3.6$, $8.2$ and $7.4$. Top-$M$ improves least on Test A
because SFT already reaches $92.0\%$ there. Because Tests A, B and C also differ in the number and share of annotated successes ($25\%$, $40\%$ and $33\%$),
these trends reflect set size and composition together.

Second, depth and the autoregressive physical reasoning trace target affect different parts of the ordering. After
SFT, the depth-only and justification-supervised variants have similar all-versus-all accuracy, $81.6\%$ and
$80.8\%$, while justification supervision is five points higher on Top-$M$, $90.6\%$ against $85.6\%$. Depth
primarily improves the global ordering. Requiring the model to generate a physical reasoning trace before its score
primarily improves discrimination involving the best candidates, with most of that Top-$M$ gain already present
after SFT. After GRPO, combining both components adds $4.1$ points over depth alone and $2.3$
over justification supervision alone on all-versus-all accuracy. The full model reaches $95.6\%$ on Top-$M$.

\begin{table}[!htbp]
\caption{Effect of rendering the physical reasoning target. All variants use the same RGB + depth inputs and the
same candidate grasp, and the final SFT+GRPO checkpoint. Ranking metrics are on Test A; LIBERO-Plus is
zero-shot success rate (\%, mean over three
seeds). The deterministic template supplies the same procedural physical content without the Qwen rewrite.}
\label{tab:trace-rendering}
\begin{center}
\small
\setlength{\tabcolsep}{5pt}
\begin{tabular}{lccc}
\toprule
\textbf{Autoregressive output target} & \textbf{All-vs-all} & \textbf{Top-$M$} & \textbf{LIBERO-Plus} \\
\midrule
Score only                                  & 82.78 & 89.81 & 87.2\,\scriptsize{$\pm$0.6} \\
Procedural template $\rightarrow$ score     & 85.68 & 94.54 & 88.1\,\scriptsize{$\pm$0.5} \\
Qwen rewrite $\rightarrow$ score            & \textbf{86.89} & \textbf{95.56} & \textbf{88.8}\,\scriptsize{$\pm$0.8} \\
\bottomrule
\end{tabular}
\end{center}
\end{table}

\paragraph{The ranking gains carry over downstream.}
Table~\ref{tab:inputs-downstream} transfers the same assessment configurations to ACoT-VLA and evaluates the
resulting policies zero-shot on LIBERO-Plus. Components that improve the assessment diagnostic also improve
downstream success, with depth the one exception noted below. RGB score-only AAPT exceeds the base VLM by $0.8$
points. Generating a physical reasoning trace
before the score increases the gain to $1.5$ points, and adding depth produces the full $1.8$-point improvement.
These results link the assessment behavior to transfer, but they do not make the output format alone sufficient.
The permuted-target control in Table~\ref{tab:objective} retains the full RGB, depth and justification-to-score
format while
destroying candidate-quality alignment, and its downstream performance falls below the base VLM.

Depth does not transfer monotonically. Added to score-only supervision it slightly reduces downstream success, from
$87.8$ to $87.2$, even though it improves global ordering on the assessment task itself. It helps only once the model
must generate a physical reasoning trace, where it adds a further $0.3$ points. Depth is also a pre-training input
only: both downstream pipelines consume RGB and language alone, so any benefit must be carried by the learned
backbone rather than by a depth channel available at policy time. We therefore read depth as sharpening the contact
geometry that the justification has to account for, rather than as an input the policy inherits. A pre-training
signal that is useful for assessment is consequently not guaranteed to be useful for control on its own.

\begin{table}[!htbp]
\caption{Sensitivity of downstream transfer to the assessment configuration. AAPT is pre-trained with each
input/output-target combination and evaluated with ACoT-VLA, zero-shot on LIBERO-Plus (\%, mean over three seeds).
``Trace $\rightarrow$ score'' means that the trace is generated autoregressively before the score; it is never an
input. Every AAPT row also receives the candidate grasp.}
\label{tab:inputs-downstream}
\begin{center}
\small
\begin{tabular}{lc}
\toprule
\textbf{Assessment pre-training configuration} & \textbf{LIBERO-Plus} \\
\midrule
Base VLM (no AAPT)            & 87.0\,\scriptsize{$\pm$0.7} \\
AAPT, RGB input; score only                 & 87.8\,\scriptsize{$\pm$0.4} \\
AAPT, RGB + depth inputs; score only         & 87.2\,\scriptsize{$\pm$0.6} \\
AAPT, RGB input; trace $\rightarrow$ score   & 88.5\,\scriptsize{$\pm$0.7} \\
AAPT, RGB + depth inputs; trace $\rightarrow$ score & \textbf{88.8}\,\scriptsize{$\pm$0.8} \\
\bottomrule
\end{tabular}
\end{center}
\end{table}

\section{Audit of the pre-training objectives}
\label{app:objective-audit}

Table~\ref{tab:objective-audit} separates exact controls from broader objective comparisons. Every row uses the same
Qwen3-VL-4B backbone and downstream ACoT-VLA recipe. AAPT and its permuted-target control are matched in data,
inputs, output format and optimization, differing only in the assignment of targets to candidates.
Collision-and-stability prediction uses the same candidate examples, model inputs and SFT stage as AAPT. Its
traces are generated with a separate prompt that is never exposed to $r^{\mathrm{gt}}$,
only to the candidate's geometry and the binary outcomes
(Figure~\ref{fig:traceprompt-control}). It replaces the scalar
assessment score with two categorical outcomes and omits GRPO. Its two
indicators are measured in the simulator for every candidate in the corpus, whereas AAPT never queries the simulator
and derives its target from annotated grasp geometry alone. The control is therefore not disadvantaged by weaker
labels: it is supervised by physical outcomes, while AAPT is supervised by a geometric proxy for them. We do not
substitute these simulated outcomes for the AAPT target because they are binary. As argued in
Section~\ref{sec:method}, binary supervision makes a near-miss indistinguishable from a gross failure, and the
graded ordering that distinguishes them is the object of study. The two rows therefore contrast a predictive
objective on true outcomes against an evaluative objective on graded proxies, rather than contrasting label quality.
The SFT-only AAPT row completes this comparison. It shares scenes, candidates, inputs, example count and SFT budget
with the collision-and-stability row, while replacing its binary outcomes and outcome-conditioned trace with a
graded trace-and-score target. This comparison avoids the confound of an additional optimization stage, although
the different target types necessarily use different output formats. The full AAPT row then measures what ranking
refinement adds on top.
Grasp-pose prediction shares the source scenes and successful grasps but does not receive failures or
candidate-conditioned targets. The VQA objectives use their own descriptive corpora and textual targets. Thus,
collision-and-stability prediction is the strongest same-data predictive control, while the other rows test
representative alternative pre-training objectives.

\begin{table}[!htbp]
\caption{Detailed audit of the pre-training objectives in Table~\ref{tab:objective}, using the same row groups and
order. ``Privileged'' geometric fields are used only by the offline trace generator to construct output targets;
they are never model inputs.}
\label{tab:objective-audit}
\begin{center}
\scriptsize
\setlength{\tabcolsep}{3pt}
\renewcommand{\arraystretch}{1.15}
\begin{tabular}{p{0.12\textwidth}p{0.23\textwidth}p{0.14\textwidth}p{0.28\textwidth}p{0.14\textwidth}}
\toprule
\textbf{Objective} & \textbf{Data and composition} & \textbf{Model input} &
\textbf{Target and supervision} & \textbf{Budget and optimization} \\
\midrule
Base VLM, no pre-training &
No additional pre-training data &
-- &
-- &
-- \\
\midrule
\multicolumn{5}{l}{\textbf{Alternative objectives}}\\
Robo2VLM &
Robo2VLM-1 teleoperation-derived robot-manipulation VQA~\citep{robo2vlm}; no success/failure structure &
RGB &
Textual answer to an embodied-VQA prompt; no AAPT geometry or physical-reasoning trace &
684K examples; 1 SFT epoch \\
\midrule
VQASynth &
Synthetic embodied VQA generated from BridgeData V2 with VQASynth~\citep{vqasynth}; no success/failure structure &
RGB &
Textual answer to a synthetic embodied-VQA prompt; no AAPT geometry or physical-reasoning trace &
500K examples; 1 SFT epoch \\
\midrule
Grasp-pose prediction &
The same HouseCat6D and GraspNet-1Billion scenes and successful grasps as AAPT; positives only &
RGB and depth; no candidate pose &
The successful grasp pose; no generated reasoning trace, failures, scores or candidate comparisons &
500K examples; 1 SFT epoch \\
\midrule
\multicolumn{5}{l}{\textbf{Matched AAPT progression}}\\
Collision + stability prediction &
Exactly the same scenes, candidates and success/failure composition as AAPT &
RGB, depth and candidate grasp &
A physical reasoning trace generated with a separate prompt that is never exposed to $r^{\mathrm{gt}}$,
only to geometry and binary outcomes (Figure~\ref{fig:traceprompt-control}),
followed by binary indicators for
environment collision and grasp stability instead of a scalar score; both indicators are measured in the simulator
for every candidate; no pairwise ranking target &
500K examples; 1 SFT epoch; no GRPO \\
\midrule
Graded SFT (Stage 1 only) &
Exactly the same scenes, candidates and success/failure composition as AAPT &
RGB, depth and candidate grasp &
Identical to AAPT: the same physical reasoning trace $\rightarrow$ \texttt{<score>} $\rightarrow$ graded scalar score.
It replaces the control's outcome-conditioned trace and binary indicators with a trace written to justify $r^{\mathrm{gt}}$
and the graded scalar score &
500K examples; 1 SFT epoch; no GRPO \\
\midrule
\rowcolor{blktwo} AAPT &
HouseCat6D and GraspNet-1Billion; successes, four $\mathrm{SE}(3)$ perturbation levels and dataset or uniform
failures; graded rewards &
RGB, depth and candidate grasp &
Qwen3-VL-8B rewrites a privileged geometric scaffold into a physical reasoning trace
$\rightarrow$ \texttt{<score>} $\rightarrow$ scalar score; privileged fields are absent from the model prompt &
500K examples; 1 SFT epoch + 500 GRPO steps \\
\midrule
\multicolumn{5}{l}{\textbf{Target-alignment control}}\\
AAPT, permuted targets &
Exactly the same scenes, candidates and success/failure composition as AAPT &
RGB, depth and candidate grasp &
The same trace $\rightarrow$ score sequence as AAPT, but each candidate receives another candidate's paired trace
and reward from the same scene. SFT uses the permuted sequence and GRPO uses the corresponding permuted reward &
500K examples; 1 SFT epoch + 500 GRPO steps \\
\bottomrule
\end{tabular}
\end{center}
\end{table}

\section{Seed-wise paired results}
\label{app:paired}

Table~\ref{tab:paired-seeds} gives the matched-seed values behind the main ACoT-VLA comparisons. Within each row,
the Base and AAPT policies share the training seed and data order, and $\Delta$ is computed before aggregation.
Summary rows report the mean and sample standard deviation across three training seeds. The standard deviation in
the final column is computed directly from the three paired differences, not from the marginal standard deviations.
Rollout episodes estimate a checkpoint's success rate, and the seven LIBERO-Plus factors are correlated evaluations
of the same checkpoints. Neither is treated as an independent training replicate.

\begin{table}[!htbp]
\caption{Seed-wise paired ACoT-VLA results. LIBERO and LIBERO-Plus report success rate (\%); CALVIN reports average
sequence length. $\Delta=\text{AAPT}-\text{Base}$. Summary rows give mean $\pm$ sample SD, including the SD of the
within-seed differences.}
\label{tab:paired-seeds}
\begin{center}
\small
\setlength{\tabcolsep}{7pt}
\begin{tabular}{lcrrr}
\toprule
\textbf{Benchmark} & \textbf{Seed} & \textbf{Base} & \textbf{AAPT} & \textbf{Paired $\Delta$} \\
\midrule
LIBERO & 1 & 97.1 & 97.8 & +0.7 \\
LIBERO & 2 & 98.3 & 98.6 & +0.3 \\
LIBERO & 3 & 98.0 & 98.2 & +0.2 \\
\textit{LIBERO summary} & -- & $97.8\pm0.6$ & $98.2\pm0.4$ & $+0.4\pm0.3$ \\
\midrule
LIBERO-Plus & 1 & 86.4 & 88.0 & +1.6 \\
LIBERO-Plus & 2 & 87.7 & 89.6 & +1.9 \\
LIBERO-Plus & 3 & 86.9 & 88.8 & +1.9 \\
\textit{LIBERO-Plus summary} & -- & $87.0\pm0.7$ & $88.8\pm0.8$ & $+1.8\pm0.2$ \\
\midrule
CALVIN & 1 & 4.35 & 4.52 & +0.17 \\
CALVIN & 2 & 4.32 & 4.53 & +0.21 \\
CALVIN & 3 & 4.41 & 4.57 & +0.16 \\
\textit{CALVIN summary} & -- & $4.36\pm0.05$ & $4.54\pm0.03$ & $+0.18\pm0.03$ \\
\bottomrule
\end{tabular}
\end{center}
\end{table}

\paragraph{Paired objective-level contrasts.}
Table~\ref{tab:paired-progression} extends the matched-seed analysis to every row of the matched AAPT progression in
Table~\ref{tab:objective}. Table~\ref{tab:paired-contrasts} reports the paired difference between consecutive rows.
Each increment is positive on all three seeds and on both benchmarks, and its paired standard deviation is well below
the marginal seed standard deviation of either arm, because matched runs share seed and data-order effects. The
three increments sum to the full paired gain over the base.

\begin{table}[!htbp]
\caption{Seed-wise results for the matched AAPT progression with ACoT-VLA. LIBERO-Plus is success rate (\%) and
CALVIN is average sequence length. Base and AAPT values are those of Table~\ref{tab:paired-seeds}.}
\label{tab:paired-progression}
\begin{center}
\small
\setlength{\tabcolsep}{6pt}
\begin{tabular}{lcrrrr}
\toprule
\textbf{Benchmark} & \textbf{Seed} & \textbf{Base} & \textbf{Collision + stability} & \textbf{Graded SFT} &
\textbf{AAPT} \\
\midrule
LIBERO-Plus & 1 & 86.4 & 87.10 & 87.55 & 88.0 \\
LIBERO-Plus & 2 & 87.7 & 88.00 & 88.75 & 89.6 \\
LIBERO-Plus & 3 & 86.9 & 87.30 & 88.10 & 88.8 \\
\textit{LIBERO-Plus summary} & -- & $87.0\pm0.7$ & $87.5\pm0.5$ & $88.1\pm0.6$ & $88.8\pm0.8$ \\
\midrule
CALVIN & 1 & 4.35 & 4.42 & 4.45 & 4.52 \\
CALVIN & 2 & 4.32 & 4.39 & 4.43 & 4.53 \\
CALVIN & 3 & 4.41 & 4.44 & 4.48 & 4.57 \\
\textit{CALVIN summary} & -- & $4.36\pm0.05$ & $4.42\pm0.03$ & $4.45\pm0.03$ & $4.54\pm0.03$ \\
\bottomrule
\end{tabular}
\end{center}
\end{table}

\begin{table}[!htbp]
\caption{Paired differences between consecutive rows of the matched AAPT progression, computed within seed (mean
$\pm$ sample SD over three seeds).}
\label{tab:paired-contrasts}
\begin{center}
\small
\begin{tabular}{lcccc}
\toprule
\textbf{Contrast} & \textbf{LIBERO-Plus $\Delta$} & \textbf{Seeds $>0$} & \textbf{CALVIN $\Delta$} &
\textbf{Seeds $>0$} \\
\midrule
Collision + stability $-$ Base & $+0.5\pm0.2$ & 3/3 & $+0.06\pm0.02$ & 3/3 \\
Graded SFT $-$ Collision + stability & $+0.7\pm0.2$ & 3/3 & $+0.04\pm0.01$ & 3/3 \\
AAPT $-$ Graded SFT & $+0.7\pm0.2$ & 3/3 & $+0.09\pm0.02$ & 3/3 \\
\midrule
AAPT $-$ Base & $+1.8\pm0.2$ & 3/3 & $+0.18\pm0.03$ & 3/3 \\
\bottomrule
\end{tabular}
\end{center}
\end{table}

\section{Additional controls}
\label{app:controls}

All controls in this section use Qwen3-VL-4B, the $500$K corpus and the ACoT-VLA recipe of
Sec.~\ref{app:downstream-protocols}, and share seeds and data order with the corresponding rows of
Table~\ref{tab:paired-progression}. Two of them add compute at different stages. \emph{Continued assessment SFT}
spends extra compute in pre-training, on the assessment corpus, and is matched to the GRPO stage.
\emph{Longer policy training} spends extra compute downstream, on the robot demonstrations, and is compared with the
whole AAPT pre-training budget.

\paragraph{Ranking refinement versus additional optimization.}
Table~\ref{tab:stage2} starts every row from the same SFT-only AAPT checkpoint. Continued assessment SFT trains on the same
data with GRPO's learning rate for the same GPU-hours as the GRPO stage. Because GPU-hours include rollout
generation, this gives continued assessment SFT more update steps than GRPO receives. Calibration-only GRPO uses the GRPO
schedule of Sec.~\ref{app:optimization} with $\alpha=0$, so its reward contains only the calibration term.
Matched continued assessment SFT adds $+0.1$ on LIBERO-Plus and $+0.01$ on CALVIN. Adding the ranking term to calibration-only
GRPO accounts for most of the stage-2 gain on both benchmarks, $+0.5$ of $+0.7$ on LIBERO-Plus and $+0.06$ of $+0.09$
on CALVIN. On CALVIN, the paired difference between full AAPT and calibration-only GRPO is $+0.06\pm0.01$ and
positive on all three seeds.

\begin{table}[!htbp]
\caption{Stage-2 controls from the same SFT-only checkpoint (mean $\pm$ SD over three seeds).}
\label{tab:stage2}
\begin{center}
\small
\begin{tabular}{lcc}
\toprule
\textbf{Stage 2} & \textbf{LIBERO-Plus} & \textbf{CALVIN} \\
\midrule
None (Graded SFT) & $88.1\pm0.6$ & $4.45\pm0.03$ \\
Continued assessment SFT, GPU-hours matched to GRPO & $88.2\pm0.5$ & $4.46\pm0.03$ \\
GRPO, calibration reward only ($\alpha=0$) & $88.3\pm0.6$ & $4.48\pm0.03$ \\
\rowcolor{blktwo} GRPO, ranking and calibration (AAPT, $\alpha=0.6$) & $\mathbf{88.8\pm0.8}$ & $\mathbf{4.54\pm0.03}$ \\
\bottomrule
\end{tabular}
\end{center}
\end{table}

\paragraph{Separating trace and score alignment.}
The permuted-target control of Table~\ref{tab:objective} breaks the alignment of the trace and the score together.
Table~\ref{tab:split-shuffle} breaks one at a time under the same data, budget and schedule. In the scores-only
shuffle each candidate keeps its own trace but receives another candidate's score from the same scene; in the
traces-only shuffle it keeps its own score but receives another candidate's trace. Corrupting either
component removes the full AAPT gain: an aligned trace with a shuffled score stays near the base, an aligned score
with a shuffled trace falls below it, and breaking both falls furthest. Because each partial shuffle retains one
corrupted target, these diagnostics establish sensitivity to joint alignment rather than the sufficiency of either
component in isolation. These results are for a single training seed and are compared with the same-seed base and
AAPT runs.

\begin{table}[!htbp]
\caption{Separate trace and score shuffles, training seed 1. Base and AAPT are the seed-1 values of
Table~\ref{tab:paired-seeds}.}
\label{tab:split-shuffle}
\begin{center}
\small
\begin{tabular}{lcccc}
\toprule
\textbf{Pre-training} & \textbf{Trace} & \textbf{Score} & \textbf{LIBERO-Plus} & \textbf{CALVIN} \\
\midrule
\rowcolor{blktwo} AAPT & aligned & aligned & $\mathbf{88.0}$ & $\mathbf{4.52}$ \\
Base, no pre-training & -- & -- & $86.4$ & $4.35$ \\
Scores-only shuffle & aligned & shuffled & $86.6$ & $4.31$ \\
Traces-only shuffle & shuffled & aligned & $85.9$ & $4.29$ \\
Full permutation & shuffled & shuffled & $81.9$ & $4.25$ \\
\bottomrule
\end{tabular}
\end{center}
\end{table}

\paragraph{Spending the pre-training compute downstream.}
AAPT pre-training, one SFT epoch over the $500$K corpus followed by $500$ GRPO steps, costs an estimated
$2.3\times10^{19}$ FLOPs, based on $6N$ training and $2N$ generation FLOPs per token for $N\approx4$B parameters. One
ACoT-VLA step costs about $6\times10^{14}$ FLOPs, so AAPT's pre-training compute corresponds to roughly $38$K
downstream steps. Table~\ref{tab:downstream-compute} trains the base-initialized policy for $30$K additional steps on
the same demonstrations, $50\%$ beyond the standard $60$K-step schedule and comparable to that estimate. In this
single-seed comparison, the extra downstream training gives smaller gains than AAPT on all three benchmarks.
Without additional demonstrations, longer policy training does not reproduce the AAPT gain on this seed.
These results are for a single training seed.

\begin{table}[!htbp]
\caption{Longer policy training of the base-initialized ACoT-VLA on the same demonstrations, training seed 1. Base and AAPT are the
seed-1 values of Table~\ref{tab:paired-seeds}.}
\label{tab:downstream-compute}
\begin{center}
\small
\begin{tabular}{lccc}
\toprule
\textbf{Initialization and schedule} & \textbf{LIBERO} & \textbf{LIBERO-Plus} & \textbf{CALVIN} \\
\midrule
Base, standard policy training ($60$K steps) & $97.1$ & $86.4$ & $4.35$ \\
Base, longer policy training ($+30$K steps) & $97.4$ & $86.7$ & $4.38$ \\
\rowcolor{blktwo} AAPT, standard policy training ($60$K steps) & $\mathbf{97.8}$ & $\mathbf{88.0}$ & $\mathbf{4.52}$ \\
\bottomrule
\end{tabular}
\end{center}
\end{table}

\section{Compute estimates}
\label{app:compute}

We estimate cost with the standard approximation of $6N$ FLOPs per training token and $2N$ per generated or
prefilled token, where $N$ is the number of model parameters ($N\approx8$B for the Qwen3-VL-8B trace renderer and
$N\approx4$B for the Qwen3-VL-4B backbone)~\cite{kaplan2020scaling}. \cref{tab:compute} lists each stage.
AAPT pre-training (SFT and GRPO) costs roughly $0.6\times$ one downstream ACoT-VLA run, or about $1\times$ with the
one-time trace rendering. Rollout generation is memory-bound, so GRPO takes a larger share of wall-clock time than of
FLOPs. AAPT is a one-time cost per backbone and is reused across benchmarks and pipelines, whereas demonstrations
have to be collected for every task.

\begin{table}[!htbp]
\centering
\footnotesize
\setlength{\tabcolsep}{3pt}
\begin{tabular}{@{}lcc@{}}
\toprule
\textbf{Stage} & \textbf{FLOPs} & \textbf{vs.\ one policy run} \\
\midrule
Candidate generation and scaffold & CPU only & negligible \\
Trace rendering (8B, one pass per example) & $\approx1.3\times10^{19}$ & $\approx0.4\times$ \\
AAPT SFT ($500$K examples, one epoch) & $\approx1.8\times10^{19}$ & $\approx0.5\times$ \\
AAPT GRPO ($500$ steps, $5$ rollouts) & $\approx0.5\times10^{19}$ & $\approx0.15\times$ \\
ACoT-VLA training ($60$K steps) & $\approx3.6\times10^{19}$ & $1.0\times$ \\
\bottomrule
\end{tabular}
\caption{Estimated compute for each stage.}
\label{tab:compute}
\end{table}

\section{Full benchmark comparisons}
\label{app:fulltables}

The main text abbreviates the published comparisons for readability. Tables~\ref{tab:libero-full},
\ref{tab:calvin-full} and \ref{tab:liberoplus-full} restore all collected methods. Our rows, evaluation protocols
and method groupings are unchanged. Published rows are reproduced from the cited papers and may differ in training
data, backbone size and policy architecture, so these tables provide context rather than controlled attribution.

\begin{table}[!htbp]
\caption{LIBERO success rate (\%), full comparison. Grouping and protocol are as in Table~\ref{tab:bench}.}
\label{tab:libero-full}
\begin{center}
\small
\begin{tabular}{lccccc}
\toprule
\textbf{Method} & \textbf{Spatial} & \textbf{Object} & \textbf{Goal} & \textbf{Long} & \textbf{Avg} \\
\midrule
\addlinespace[1pt]\multicolumn{6}{l}{\textbf{Direct policies}}\\\addlinespace[1pt]
Diffusion Policy~\citep{diffusion-policy} & 78.3 & 92.5 & 68.3 & 50.5 & 72.4 \\
Octo~\citep{octo} & 78.9 & 85.7 & 84.6 & 51.1 & 75.1 \\
OpenVLA~\citep{openvla} & 84.9 & 88.4 & 79.2 & 53.7 & 76.5 \\
UniAct~\citep{uniact} & 77.0 & 87.0 & 77.0 & 70.0 & 77.8 \\
SpatialVLA~\citep{spatialvla} & 88.2 & 89.9 & 78.6 & 55.5 & 78.1 \\
DiT Policy~\citep{ditpolicy} & 84.2 & 96.3 & 85.4 & 63.8 & 82.4 \\
ThinkAct~\citep{thinkact} & 88.3 & 91.4 & 87.1 & 70.9 & 84.4 \\
$\pi_0$-FAST~\citep{pi0} & 96.4 & 96.8 & 88.6 & 60.2 & 85.5 \\
FPC-VLA~\citep{fpcvla} & 87.0 & 92.0 & 86.2 & 82.2 & 86.9 \\
SmolVLA~\citep{smolvla} & 93.0 & 94.0 & 91.0 & 77.0 & 88.8 \\
GR00T-N1~\citep{groot-n1} & 94.4 & 97.6 & 93.0 & 90.6 & 93.9 \\
$\pi_0$~\citep{pi0} & 96.8 & 98.8 & 95.8 & 85.2 & 94.2 \\
VLA-0~\citep{vla0} & 97.0 & 97.8 & 96.2 & 87.6 & 94.7 \\
GO-1~\citep{go1} & 96.2 & 97.8 & 96.0 & 89.2 & 94.8 \\
Discrete Diffusion VLA~\citep{ddvla} & 97.2 & 98.6 & 97.4 & 92.0 & 96.3 \\
MemoryVLA~\citep{memoryvla} & 98.4 & 98.4 & 96.4 & 93.4 & 96.7 \\
$\pi_{0.5}$~\citep{pi05} & 98.8 & 98.2 & 98.0 & 92.4 & 96.9 \\
OpenVLA-OFT~\citep{openvla-oft} & 97.6 & 98.4 & 97.9 & 94.5 & 97.1 \\
VLA-Adapter~\citep{vla-adapter} & 97.8 & 99.2 & 97.2 & 95.0 & 97.3 \\
TurboVLA~\citep{turbovla} & 99.2 & 99.8 & 97.4 & 94.2 & 97.7 \\
ACoT-VLA~\citep{acot-vla} & 98.6 & 99.0 & 99.4 & 97.0 & 98.5 \\
\midrule
\addlinespace[1pt]\multicolumn{6}{l}{\textbf{Prediction \& world models}}\\\addlinespace[1pt]
CoT-VLA~\citep{cot-vla} & 87.5 & 91.6 & 87.6 & 69.0 & 81.1 \\
WorldVLA~\citep{worldvla} & 87.6 & 96.2 & 83.4 & 60.0 & 81.8 \\
Seer~\citep{seer} & -- & -- & -- & 87.7 & 87.7 \\
FlowVLA~\citep{flowvla} & 93.2 & 95.0 & 91.6 & 72.6 & 88.1 \\
DreamVLA~\citep{dreamvla} & 97.5 & 94.0 & 89.5 & 89.5 & 92.6 \\
UD-VLA~\citep{udvla} & 94.1 & 95.7 & 91.2 & 89.6 & 92.7 \\
RynnVLA-002~\citep{rynnvla} & 94.2 & 96.4 & 94.6 & 87.6 & 93.3 \\
UniVLA~\citep{univla} & 95.4 & 98.8 & 93.5 & 94.0 & 95.5 \\
F1~\citep{f1vla} & 98.2 & 97.8 & 95.4 & 91.3 & 95.7 \\
GE-Act~\citep{geact} & 98.2 & 97.6 & 95.8 & 94.4 & 96.5 \\
MMaDA-VLA~\citep{mmada-vla} & 98.8 & 99.8 & 98.0 & 95.2 & 98.0 \\
\midrule
\addlinespace[1pt]\multicolumn{6}{l}{\textbf{Ours}, on Qwen3-VL-4B}\\\addlinespace[1pt]
\rowcolor{blkone} VLM4VLA        & 80.2 & 87.2 & 72.4 & 44.4 & 71.1\,\scriptsize{$\pm$2.4} \\
\rowcolor{blkone} AAPT + VLM4VLA & 84.0 & 90.2 & 77.6 & 60.8 & \textbf{78.2}\,\scriptsize{$\pm$2.1} \\
\rowcolor{blktwo} ACoT-VLA        & 98.4 & 98.6 & 98.4 & 95.8 & 97.8\,\scriptsize{$\pm$0.6} \\
\rowcolor{blktwo} AAPT + ACoT-VLA & 98.8 & 99.4 & 98.8 & 96.0 & \textbf{98.2}\,\scriptsize{$\pm$0.4} \\
\bottomrule
\end{tabular}
\end{center}
\end{table}

\begin{table}[!htbp]
\caption{CALVIN ABC$\rightarrow$D, full comparison. Grouping and columns are as in Table~\ref{tab:bench}.}
\label{tab:calvin-full}
\begin{center}
\small
\begin{tabular}{lcccccc}
\toprule
\textbf{Method} & \textbf{1/5} & \textbf{2/5} & \textbf{3/5} & \textbf{4/5} & \textbf{5/5} & \textbf{Avg.\ Len.} \\
\midrule
\addlinespace[1pt]\multicolumn{7}{l}{\textbf{Direct policies}}\\\addlinespace[1pt]
Diffusion Policy~\citep{diffusion-policy} & 40.2 & 12.3 & 3.2 & 1.3 & 0.0 & 0.56 \\
RT-1~\citep{rt1} & 53.3 & 22.2 & 9.4 & 3.8 & 1.3 & 0.90 \\
VLAS~\citep{vlas} & 87.2 & 64.2 & 40.9 & 28.1 & 19.6 & 2.40 \\
RoboFlamingo~\citep{roboflamingo} & 82.4 & 61.9 & 46.6 & 33.1 & 23.5 & 2.47 \\
DeeR-VLA~\citep{deervla} & 84.8 & 72.3 & 54.9 & 44.6 & 33.5 & 2.90 \\
OpenVLA~\citep{openvla} & 91.3 & 77.8 & 62.0 & 52.1 & 43.5 & 3.27 \\
$\pi_0$~\citep{pi0} & 96.8 & 85.0 & 76.7 & 68.1 & 59.9 & 3.92 \\
LLaDA-VLA~\citep{lladavla} & 95.6 & 87.7 & 79.5 & 73.9 & 64.5 & 4.01 \\
OpenHelix~\citep{openhelix} & 97.1 & 91.4 & 82.8 & 72.6 & 64.1 & 4.08 \\
OpenVLA-OFT~\citep{openvla-oft} & 96.3 & 89.1 & 82.4 & 75.8 & 66.5 & 4.10 \\
RoboVLMs~\citep{robovlms} & 98.0 & 93.6 & 85.2 & 77.9 & 70.4 & 4.25 \\
VLA-Adapter~\citep{vla-adapter} & 99.1 & 94.6 & 88.8 & 82.8 & 76.5 & 4.42 \\
\midrule
\addlinespace[1pt]\multicolumn{7}{l}{\textbf{Prediction \& world models}}\\\addlinespace[1pt]
SuSIE~\citep{susie} & 87.0 & 69.0 & 49.0 & 38.0 & 26.0 & 2.69 \\
GEVRM~\citep{gevrm} & 92.0 & 70.0 & 54.0 & 41.0 & 26.0 & 2.83 \\
GR-1~\citep{gr1} & 85.4 & 71.2 & 59.6 & 49.7 & 40.1 & 3.06 \\
CLOVER~\citep{clover} & 96.0 & 83.5 & 70.8 & 57.5 & 45.4 & 3.53 \\
UniVLA~\citep{univla13} & 95.5 & 85.8 & 75.4 & 66.9 & 56.5 & 3.80 \\
ReconVLA~\citep{reconvla} & 95.6 & 87.6 & 76.9 & 69.3 & 64.1 & 3.95 \\
UP-VLA~\citep{upvla} & 92.8 & 86.5 & 81.5 & 76.9 & 69.9 & 4.08 \\
Seer~\citep{seer} & 96.3 & 91.6 & 86.1 & 80.3 & 74.0 & 4.28 \\
VPP~\citep{vpp} & 95.7 & 91.2 & 86.3 & 81.0 & 75.0 & 4.33 \\
UniVLA~\citep{univla} & 98.9 & 94.8 & 89.0 & 82.6 & 75.1 & 4.41 \\
DreamVLA~\citep{dreamvla} & 98.2 & 94.6 & 89.5 & 83.4 & 78.1 & 4.44 \\
MMaDA-VLA~\citep{mmada-vla} & 99.8 & 98.6 & 96.3 & 93.5 & 89.7 & 4.78 \\
\midrule
\addlinespace[1pt]\multicolumn{7}{l}{\textbf{Ours}, on Qwen3-VL-4B}\\\addlinespace[1pt]
\rowcolor{blkone} VLM4VLA         & 93.3 & 85.7 & 79.0 & 71.9 & 64.4 & 3.94\,\scriptsize{$\pm$0.08} \\
\rowcolor{blkone} AAPT + VLM4VLA  & 94.4 & 88.1 & 83.2 & 77.4 & 70.9 & \textbf{4.14}\,\scriptsize{$\pm$0.06} \\
\rowcolor{blktwo} ACoT-VLA        & 96.1 & 92.1 & 87.3 & 82.8 & 77.6 & 4.36\,\scriptsize{$\pm$0.05} \\
\rowcolor{blktwo} AAPT + ACoT-VLA & 97.2 & 94.9 & 90.3 & 86.7 & 84.8 & \textbf{4.54}\,\scriptsize{$\pm$0.03} \\
\bottomrule
\end{tabular}
\end{center}
\end{table}

\begin{table}[!htbp]
\caption{Zero-shot LIBERO-Plus success rate (\%), full comparison. Seven held-out perturbation factors, evaluated without fine-tuning on the LIBERO checkpoints of Table~\ref{tab:bench};
$^{*}$ reproduced from released checkpoints by~\citet{acot-vla}.}
\label{tab:liberoplus-full}
\begin{center}
\small
\setlength{\tabcolsep}{2pt}
\begin{tabular}{lcccccccc}
\toprule
\textbf{Method} & \textbf{Camera} & \textbf{Robot} & \textbf{Lang.} & \textbf{Light} & \textbf{Backgr.} & \textbf{Noise} & \textbf{Layout} & \textbf{Avg} \\
\midrule
OpenVLA~\citep{openvla} & 0.8 & 3.5 & 23.0 & 8.1 & 34.8 & 15.2 & 28.5 & 16.3 \\
WorldVLA~\citep{worldvla} & 0.1 & 27.9 & 41.6 & 43.7 & 17.1 & 10.9 & 38.0 & 25.6 \\
NORA~\citep{nora} & 2.2 & 37.0 & 65.1 & 45.7 & 58.6 & 12.8 & 62.1 & 40.5 \\
UniVLA~\citep{univla} & 1.8 & 46.2 & 69.6 & 69.0 & 81.0 & 21.2 & 31.9 & 45.8 \\
$\pi_0$-FAST~\citep{pi0} & 65.1 & 21.6 & 61.0 & 73.2 & 73.2 & 74.4 & 68.8 & 62.5 \\
RIPT-VLA~\citep{riptvla} & 55.2 & 31.2 & 77.6 & 88.4 & 91.6 & 73.5 & 74.2 & 70.2 \\
$\pi_0^{*}$~\citep{pi0} & 61.0 & 40.8 & 63.5 & 89.3 & 84.1 & 80.1 & 76.4 & 70.7 \\
OpenVLA-OFT~\citep{openvla-oft} & 56.4 & 31.9 & 79.5 & 88.7 & 93.3 & 75.8 & 74.2 & 71.4 \\
$\pi_{0.5}^{*}$~\citep{pi05} & 75.8 & 79.4 & 83.3 & 95.5 & 95.0 & 89.6 & 87.0 & 86.5 \\
ACoT-VLA~\citep{acot-vla} & 72.6 & 82.6 & 87.5 & 97.7 & 96.5 & 87.8 & 88.1 & 87.5 \\
\midrule
\addlinespace[1pt]\multicolumn{9}{l}{\textbf{Ours}, on Qwen3-VL-4B}\\\addlinespace[1pt]
\rowcolor{blktwo} ACoT-VLA        & 72.4 & 82.2 & 88.0 & 96.1 & 95.4 & 87.2 & 87.9 & 87.0\,\scriptsize{$\pm$0.7} \\
\rowcolor{blktwo} AAPT + ACoT-VLA & \textbf{75.4} & \textbf{83.4} & \textbf{88.8} & \textbf{97.5} & \textbf{97.2} & \textbf{89.6} & \textbf{89.5} & \textbf{88.8}\,\scriptsize{$\pm$0.8} \\
\bottomrule
\end{tabular}
\end{center}
\end{table}

\section{Conditional per-step reliability on CALVIN}
\label{app:conditional}

The columns of Table~\ref{tab:bench} are cumulative and report the probability of completing at least $k$
instructions. They therefore decay multiplicatively even when conditional per-step reliability is constant.
Table~\ref{tab:conditional} removes this compounding. For step one, it reports the original one-step success rate.
For each later step $k$, it reports
$P(\text{complete }k\mid\text{completed }k{-}1)=C_k/C_{k-1}$, where $C_k$ is the cumulative rate in
Table~\ref{tab:bench}. Published-method values are computed from their reported cumulative columns in the same way.

\begin{table}[!htbp]
\caption{Conditional per-step success on CALVIN ABC$\rightarrow$D. $\Delta$ is in percentage points.}
\label{tab:conditional}
\begin{center}
\small
\begin{tabular}{lccccccccc}
\toprule
& \multicolumn{3}{c}{\textbf{VLM4VLA}} & \multicolumn{3}{c}{\textbf{ACoT-VLA}} & \multicolumn{3}{c}{\textbf{Published}} \\
\cmidrule(lr){2-4}\cmidrule(lr){5-7}\cmidrule(lr){8-10}
\textbf{Step} & Base & + AAPT & $\Delta$ & Base & + AAPT & $\Delta$ & Seer & VLA-Adapter & MMaDA-VLA \\
\midrule
1 & 0.933 & 0.944 & +1.1 & 0.961 & 0.972 & +1.1 & 0.963 & 0.991 & 0.998 \\
2 & 0.919 & 0.933 & +1.5 & 0.958 & 0.976 & +1.8 & 0.951 & 0.955 & 0.988 \\
3 & 0.922 & 0.944 & +2.3 & 0.948 & 0.952 & +0.4 & 0.940 & 0.939 & 0.977 \\
4 & 0.910 & 0.930 & +2.0 & 0.948 & 0.960 & +1.2 & 0.933 & 0.932 & 0.971 \\
5 & 0.896 & 0.916 & +2.0 & 0.937 & 0.978 & +4.1 & 0.922 & 0.924 & 0.959 \\
\bottomrule
\end{tabular}
\end{center}
\end{table}

Conditional reliability varies much less across chain position than the cumulative columns suggest, including for
the strongest published methods. In our pipelines, however, the AAPT advantage is larger toward the end of the
chain. With VLM4VLA, the conditional gain rises from $1.1$ points at step one to $2.0$ points at step five, with a
maximum of $2.3$ at step three. With ACoT-VLA, the gain stays within $1.8$ points through step four and reaches
$4.1$ points at the final step. This is also where the base model has its lowest conditional reliability, $0.937$.
The widening cumulative AAPT gap is therefore not explained only by repeatedly compounding a constant advantage.

\section{Learning dynamics with scarce demonstrations}
\label{app:lowdata}

\cref{tab:lowdata} evaluates the ACoT-VLA runs trained on $10\%$ of the LIBERO demonstrations at intermediate
checkpoints, zero-shot on LIBERO-Plus. AAPT is ahead at every checkpoint. The gap is largest early, so part of the
advantage is faster learning, and AAPT at $40$K steps already exceeds the base at the end of training. The gap then
holds steady rather than closing while both curves flatten, so AAPT also reaches a higher level under the same
schedule. With all demonstrations removed but $10\%$, AAPT's final $70.0$ matches what the base reaches with roughly
$16$ to $17\%$ of the demonstrations, by linear interpolation of \cref{fig:gain}.

\begin{table}[!htbp]
\centering
\footnotesize
\begin{tabular}{@{}lccc@{}}
\toprule
\textbf{Training steps} & \textbf{Base} & \textbf{AAPT} & \textbf{Gap} \\
\midrule
$20$K & $49.8$ & $58.5$ & $+8.7$ \\
$40$K & $60.7$ & $67.1$ & $+6.4$ \\
$60$K (final) & $63.1$ & $70.0$ & $+6.9$ \\
\bottomrule
\end{tabular}
\caption{Zero-shot LIBERO-Plus success (\%) of ACoT-VLA trained on $10\%$ of the LIBERO demonstrations.}
\label{tab:lowdata}
\end{table}

\section{Does the assessment use the scene?}
\label{app:audit}

\paragraph{Scene controls.} We evaluate the final AAPT checkpoint on Tests A, B and C under three input
manipulations, with no retraining (\cref{tab:scene-controls}). \emph{Candidate only} removes the scene images and keeps
the candidate grasp text (pose and width). \emph{Mismatched scene} pairs each candidate grasp with the RGB and depth images of another
held-out scene while keeping the original target. \emph{Consistent mirror} flips RGB and depth horizontally and
mirrors the pose to match, negating the camera-frame $x$ coordinate and reflecting the rotation as
$\mathbf{R}'=\mathbf{S}\mathbf{R}\mathbf{S}$ with $\mathbf{S}=\mathrm{diag}(-1,1,1)$. The gripper width is unchanged,
and the parallel-jaw gripper is symmetric under this reflection, so the target is unchanged. Without the images, accuracy falls to a few points above
the untrained backbone. A mismatched scene lowers it below the candidate-only level, so the model uses the scene it is
given, and a consistent mirror preserves most of the accuracy.

\begin{table}[!htbp]
\centering
\footnotesize
\begin{tabular}{@{}lccc@{}}
\toprule
\textbf{Input to the AAPT checkpoint} & \textbf{Test A} & \textbf{Test B} & \textbf{Test C} \\
\midrule
RGB, depth and candidate grasp (original) & $86.9$ & $80.1$ & $74.1$ \\
Candidate grasp only & $62.5$ & $63.1$ & $59.8$ \\
Pose with another scene's images & $57.2$ & -- & -- \\
Images and pose mirrored together & $87.4$ & $78.6$ & $69.8$ \\
\midrule
Untrained base backbone (original input) & $55.8$ & $57.3$ & $53.9$ \\
Chance & $50.0$ & $50.0$ & $50.0$ \\
\bottomrule
\end{tabular}
\caption{All-versus-all ranking accuracy (\%) under scene controls. The mismatched-scene control was run on
Test A only.}
\label{tab:scene-controls}
\end{table}

\paragraph{Candidate-only AAPT.} To test whether the scene matters for transfer, we train AAPT on the same candidates and
targets with the scene images removed from the prompt, using the same SFT and GRPO schedule, and initialize ACoT-VLA
from it. On training seed 1 it reaches $86.2$ on LIBERO-Plus and $4.37$ on CALVIN, against $86.4$ and $4.35$ for the
same-seed base and $88.0$ and $4.52$ for full AAPT (\cref{tab:seed1}). Learning to rank candidates from the grasp text
alone does not produce a better initialization, so the transfer depends on assessment conditioned on the scene.

\paragraph{Trace audit.}

\paragraph{Sample.} We draw $100$ candidates from Test A ($200$ held-out scenes, scene-disjoint from training),
stratified by $r^{\mathrm{gt}}$ into four quality bins ($30$, $20$, $20$ and $30$ candidates from highest to lowest). For each, the trained
AAPT model generates a trace from RGB, depth and the candidate grasp, with no
scaffold, mesh or $\rgt$ in the prompt.

\paragraph{Judging.} An LLM judge compares each generated trace with the ground-truth procedural scaffold of the same
candidate, field by field, and we manually validated all $100$ judgments. A field counts as a match if it is stated
consistently with the scaffold within tolerance, and as a contradiction if it is stated with a wrong value or a wrong
descriptor. A trace that omits a field counts as neither, so omission is $100$ minus match minus contradiction. A
number within tolerance that contradicts a stated relation counts as an error. An unsupported claim is a physical
statement with no basis in the scaffold, and paraphrase or reordering does not count. \cref{tab:audit} in the main
paper gives the field-level results, and \cref{tab:audit-unsupported} summarizes unsupported claims. The gripper
width is part of the VLM input, so the trace's statement of the opening width is a restatement rather than an
inference. It matches within $\pm0.3$\,cm, with its relation to the object's thinnest axis preserved, in $70\%$ of
traces and is not counted as evidence of geometric recovery. Gripper fit is counted, because judging whether the
given opening suits the object requires the object's geometry.

\begin{table}[!htbp]
\centering
\footnotesize
\begin{tabular}{@{}lc@{}}
\toprule
\textbf{Measure} & \textbf{Value} \\
\midrule
Traces with at least one unsupported physical claim & $19\%$ \\
Unsupported claims per trace (mean) & $1.24$ \\
Most common types & unspecified part, obstacle, friction \\
Internally contradictory traces & $2\%$ \\
\bottomrule
\end{tabular}
\caption{Unsupported claims in $100$ AAPT-generated traces.}
\label{tab:audit-unsupported}
\end{table}

\section{Backbone and corpus scaling}
\label{app:scale}
\label{app:corpus}

\begin{figure}[H]
\centering
\begin{minipage}[b]{0.36\linewidth}
\centering
\includegraphics[width=\linewidth]{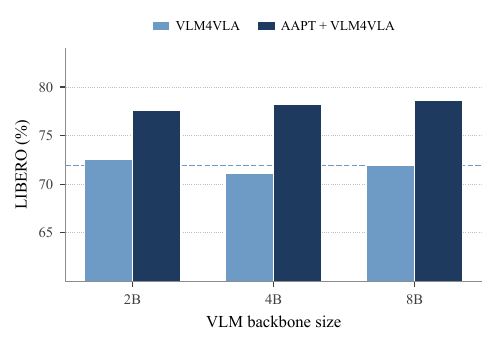}
\caption{Backbone size with VLM4VLA. The dashed line is the base VLM at 8B.}
\label{fig:backbone}
\end{minipage}\hfill
\begin{minipage}[b]{0.6\linewidth}
\centering
\includegraphics[width=\linewidth]{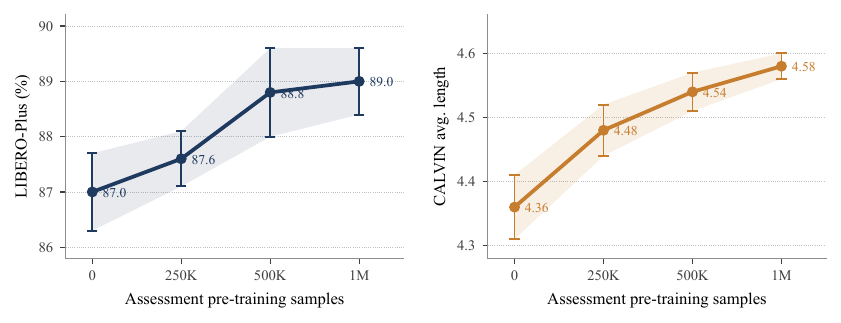}
\caption{Downstream performance versus assessment-corpus size with ACoT-VLA.}
\label{fig:corpus}
\end{minipage}
\end{figure}

\paragraph{Backbone size.} The sweep in \cref{fig:backbone} uses VLM4VLA because this minimal pipeline adds the
fewest learned components. Across the tested 2B to 8B range, increasing backbone size leaves LIBERO success nearly
flat, and AAPT with the 2B backbone already exceeds the base 8B backbone. Within this range the pre-training
objective has a larger effect than model size, although this does not imply that capacity is irrelevant beyond it.

\paragraph{Corpus size.} \cref{fig:corpus} scales AAPT to $1$M examples with ACoT-VLA. Performance improves at every
tested size on both benchmarks, and the increments shrink, with the final step to $1$M adding $+0.2$ on LIBERO-Plus
and $+0.04$ on CALVIN. The larger corpora resample candidates from the same assets, $194$ HouseCat6D instances and
$88$ GraspNet-1Billion objects, so the trend supports continued gains within the tested range but does not
establish scaling in object diversity.

\end{document}